\documentclass[11pt]{article}

\usepackage[final]{acl}
\usepackage{times}
\usepackage{cleveref}
\usepackage{latexsym}
\usepackage[dvipsnames,table]{xcolor}
\usepackage{bm}
\usepackage[hypcap=false, justification=justified]{caption}

\usepackage{float}
\usepackage[T2A,T5,T1]{fontenc}

\usepackage[utf8]{inputenc}

\usepackage{microtype}

\usepackage{inconsolata}

\usepackage{graphicx}
\usepackage{booktabs}

\usepackage{listings}
\newcommand{\epigraph}[2]{%
  \begin{quotation}
    {\itshape #1}\par\medskip
    \hfill -- #2
  \end{quotation}%
}

\title{Adopt $\neq$ Adapt:\\Longitudinal Analyses of LLM Conversations in the Wild}

\author{Rebecca M.\ M.\ Hicke\thanks{Work performed during a Microsoft Research internship.}\\
  Cornell University \\
  \texttt{rmh327@cornell.edu} \\\And
  Kiran Tomlinson \\
Microsoft Research \\
  \texttt{kitomlinson@microsoft.com} \\}

\begin{document}
\maketitle
\begin{abstract}
Although a growing body of research has begun to describe user--LLM interactions, the picture it paints is largely \textit{static}; little is known about how individual users change their behavior over time.
To address this gap, we analyze the conversational trajectories of $\sim$12,000 randomly sampled Microsoft Bing Copilot users and compare these with data from WildChat-4.8M. While the Copilot data contains significant population-level trends, we find that trends in individual user trajectories are much weaker; user habits prove to be overwhelmingly \emph{sticky}. We also find stark differences between users of different activity levels: more active users have more successful conversations and use the LLM for more complex and professionally oriented tasks. Some user trends also appear in WildChat-4.8M, but we find evidence that this dataset is significantly skewed towards highly proficient ``power'' users. Ultimately, our results suggest that existing user behavior is difficult to change and demonstrate the extent of user heterogeneity. Our comparison between datasets highlights that WildChat does not represent typical user--AI interactions, an important caveat for downstream uses of the data.
\end{abstract}

\epigraph{``The end of all our exploring will be to arrive where we started and know the place for the first time.''}{T.\ S.\ Eliot, \emph{Little Gidding}, 1942}

\section{Introduction}

A significant body of research has recently emerged studying multiple dimensions of user--LLM interactions, ranging from task complexity~\cite{suri2024complex} to user intents~\cite{handa2025economic,chatterji2025how}.  
This literature paints a clear picture of overall LLM usage; it helps system developers improve user experiences and provides guidance to users on how to productively interact with LLMs. 
However, this picture of user--LLM interactions is \textit{static} --- little is known about how individual users adapt over time.
Temporal analyses of LLM usage have primarily shown population-level trends~\cite{chatterji2025how} or cohort differences~\cite{anthropic2026aeiv5}, which do not necessarily reflect behavioral changes of individuals. Only recently have individual user trajectories emerged as an important object of study~\cite{zhu-etal-2026-show,zhu2026priming}.

LLM chats are highly unstructured environments, which often provide limited guidance on how to use the tool effectively. This places significant burden on users to discover their preferred prompting approaches and uses of AI. By studying users' behavioral trajectories, we can explore how they learn about and adapt to LLM chats and surface directions for improving long term user utility.

We consider several research questions towards this end. Primarily, we are interested in whether individuals adapt after they adopt AI. Do users change the types of tasks they perform with AI over time? Do they learn to use LLMs in more complex and successful ways? To further contextualize user-level findings, we also examine how individual-level adaptations compare to more commonly studied population-level trends and to the heterogeneity across users.

We explore these questions by analyzing six months of conversational trajectories from $\sim$12,000 randomly sampled users of Microsoft Bing Copilot active during
2024, alongside population-level daily conversation samples ($\sim$1M conversations).
We stratify the sampled users by activity level (the number of days they use Bing Copilot during our study period) and then explore five dimensions of their LLM usage: usage intensity, linguistic complexity, task completion, task intent, and conversational domain. 
These dimensions largely describe coarse features of user--LLM interactions; nonetheless, they allow us to begin describing and differentiating user behavior.
We characterize population-level trends over the study period, individual user evolution, and variation across users of different activity levels. 

We supplement this analysis with an examination of the public WildChat-4.8M dataset~\cite{zhao2024wildchat,deng2024wildvis}. 
For this data, we use hashed IP addresses as a proxy for individual users and then replicate the analyses performed on the Bing Copilot data.
Using a public dataset alongside private data provides additional transparency into our methods (e.g., allowing us to release code and classifications). 
It also allows us to compare users of a mainstream consumer LLM with the public WildChat data, which has been used extensively to understand LLM user behavior~\cite{mireshghallah2024trust}, fine-tune models~\cite{shi2025wildfeedback}, and derive benchmarks~\cite{lin2025wildbench}, among many other downstream applications.

Motivating our stratified sample, we find that frequent ``power'' users of Bing Copilot differ significantly from less frequent users along every dimension: they are more active, write more linguistically complex messages, are more likely to have completed conversations, and use the LLM for more complex and professionally-oriented tasks.
A natural assumption might be that these trends are the result of individual users learning over time.
However, our longitudinal analysis largely refutes this;
someone who will become a high-activity user interacts very differently with the LLM \emph{even early on in their trajectory}. In fact, we find that individual users change their behavior very little: user habits are \textit{sticky}.

In contrast, we find that Bing Copilot users as a \emph{group} do change over time, often towards behavior characteristic of high activity-level users. 
Again, longitudinal user data demonstrates that these changes are not caused by learning at the individual level, but are instead driven by new users who differ significantly from earlier adopters.
Finally, although our longitudinal analysis largely surfaces the stickiness of user habits, we do find some small scale changes over Bing Copilot user trajectories, including a 6--10\% increase in message complexity, a small 0--2\% increase in task completion, and a shift from exploration to exploitation. Yet, the corresponding population-level changes are  about $3\times$ larger over the same period and differences across user activity levels are $10\times$ larger. 

Some of these findings are reflected in WildChat: more active WildChat users complete more tasks and follow some similar population-level trends. However, the difference between low- and high-activity users in WildChat is far less pronounced. Our data suggest the population represented in WildChat is extremely unusual, with an over-representation of power users (likely driven by the Hugging Face front-end) and a large fraction of API-like usage rather than natural conversations. 
We caution that individual-level WildChat results may be affected by the imprecise method of identifying users by hashed IP, but population-level differences are unaffected and striking.

These findings have several implications. First, the stickiness of habits means that users may not discover more useful and successful LLM tasks through natural exploration, indicating a need for proactive interventions. Second, population-level trends toward the behavior of high-activity users suggest that learning occurs at an aggregate level, even if user rigidity inhibits it at an individual level. Third, the stark differences between users of different activity levels highlight the importance of recognizing individual heterogeneity in future work. Finally, the differences between the Bing Copilot and WildChat-4.8M user pools suggest that downstream uses of WildChat data should be cautiously implemented; WildChat-derived artifacts may not generalize to the average LLM chat user. To support further work in this area, our code  and classifications for WildChat-4.8M are available at \url{https://github.com/microsoft/adopt-adapt-wildchat}.

\section{Related Work}
A great deal of existing literature investigates how people use LLMs in the wild,
focusing on classifying user intents, conversational domains, and linguistic features of messages~\cite{handa2025economic,tomlinson2025working,chatterji2025how,chatterji2026organization,costa2026public,ouyang2023shifted,trippas2024users,zhao2024wildchat,shah2025using,suri2024complex,tamkin2024clio}. 
Some analyses explore population-level temporal trends~\cite{chatterji2025how} or differences between users of different tenure (days since signup)~\cite{anthropic2026aeiv5}. 
Unlike these works, a key portion of our analysis is at the user level (aggregated for privacy), allowing us to identify how individual behavior evolves over time.

Most existing user-level temporal analyses follow a small set of users and track how their attitudes towards and usage of AI change~\cite{skjuve2022longitudinal,skjuve2023longitudinal,long2024not,chandra2025longitudinal,ammari2025students}. However, one notable exception is a concurrent exploration of user trajectories in WildChat~\cite{zhu-etal-2026-show,zhu2026priming}, which also uses hashed IP addresses as a proxy for user identity. \citet{zhu-etal-2026-show} find that users' conversation-starting messages become more similar and stable over time, supporting our finding that user behavior is persistent. In follow-up work, \citet{zhu2026priming} additionally find that user word choice is habitual. They conclude that patterns of user expression form in only the first 3--5 conversations, and like us, advocate for user trajectories as a critical unit of analysis. Their analyses of linguistic content complement our focus on intents, message complexity, and task completion. Although we demonstrate, through comparison to the Bing Copilot data, that WildChat results should be interpreted with caution, it is encouraging that we nevertheless come to similar conclusions about the centrality of habits.

\section{Data}
\subsection{Bing Copilot}

Our main dataset consists of all conversations with Microsoft Bing Copilot in an English-language interface between January 1, 2024 and September 30, 2024.\footnote{Likely bot content is removed.} Our use of human data was approved by Microsoft Research Ethics, Privacy, and Compliance review. Personally identifying information (e.g., email addresses and financial details) was automatically scrubbed from all conversations before analysis and unique users are identified only with anonymized ids. Additionally, we report all metrics as aggregates over more than 200 users. All data was stored and classified in secure computing environments. See \Cref{app:ethics} for additional discussion of data ethics and privacy.

We aim to examine the entire trajectory of a user's interactions with Bing Copilot, from their first conversation onwards. However, the dataset does not label users' first conversations. As a proxy, we exclude all users who have conversations in the first three months of the data (before April 1, 2024), creating a filtered version of the dataset. We expect that most users who did not interact with Bing Copilot for three months before appearing in the dataset will be new.

We then take two subsamples of the filtered dataset. The first is meant to represent overall population-level user behavior during the studied time period (April 1 to September 30, 2024). To create this subsample, we randomly select $\sim$1,000 conversations from each day of the specified time period, producing a dataset of 796,838 LLM and user messages from 175,061 conversations. This we refer to as the \emph{population dataset}.

The second subsample consists of full user trajectories for a sample of users stratified by activity level, or the number of days on which the user had a conversation with Bing Copilot. 
Activity level follows a power-law distribution with exponential cutoff~\cite{clauset2009power}, with more users active on fewer days and a heavy tail. We thus categorize users into three activity-level classes: users active 1--10 days (low), active 11--25 days (middle), and active 26+ days (high).
In order to study data from users of all activity levels, we randomly sample $\sim$250 users active $1,2,...,49,50+$ days from the filtered dataset and include all of their conversations during the study period in our sample. The resulting dataset contains 812,650 conversations from 11,905 users consisting cumulatively of 4,879,568 messages. We call this the \emph{user dataset}.

\subsection{WildChat-4.8M}
Because the Bing Copilot data cannot be released publicly, we replicate our analyses on the public WildChat-4.8M dataset~\cite{zhao2024wildchat,deng2024wildvis}.\footnote{Available at \url{https://huggingface.co/datasets/allenai/WildChat-4.8M} under an ODC-By license.} WildChat was gathered by providing users with free access to GPT models through Hugging Face Spaces applications in exchange for their agreement to share data. In addition to full conversation texts, WildChat provides a hash of each user's IP address and the user's country and state. We use these fields to link conversations that may involve the same user. To remove data likely coming from shared networks, we filter out hashed IPs associated with more than three unique countries, states, or languages, and hashed IPs with more than 161 conversations (the largest number of conversations associated with at least 10 hashed IPs). This removes 3,099 hashed IPs and 677k conversations. 

We are left with 2,522,330 conversations from 1,830,631 hashed IPs that occurred between March 9, 2023 and July 31, 2025.
For simplicity, we refer to hashed IPs as ``users'' for the remainder of the paper. However, we note that this method of identifying individual users is far less reliable than that of the Bing Copilot data, where users are identified by a combination of account logins and cookies. We therefore interpret user-level results from WildChat with caution. We again split users into low, middle, and high activity groups (active 1--10, 11--25, 26+ days). WildChat's higher active day groups have far fewer users, since they are not derived from a stratified sample as in Bing Copilot.

The WildChat-4.8M dataset displays significant irregularities later in the data period, with huge spikes in the number of daily conversations and unique IPs (\Cref{app:full-wildchat}, \Cref{fig:wildchat-full-counts}). We find this is driven by API-like usage; that is, a large number of prompts with identical templates performing tasks like translation and named entity recognition (\Cref{app:full-wildchat}, \Cref{fig:templated-metrics} and \Cref{tab:templates}). To avoid the bulk of this non-conversational activity while keeping the data close to intact, the paper only considers conversations in WildChat-4.8M before September 2024. Full versions of all figures with data after this cutoff can be found in \Cref{app:full-wildchat}, alongside versions with all ``templated'' conversations removed.

\section{Methods}
\subsection{Syntactic Features}

We calculate several syntactic properties of the user--LLM conversations: the number of user messages sent per conversation, the average user sentence length per conversation,\footnote{A proxy for linguistic complexity~\citep{dale1934study, gray1935makes, flesch1948new}; calculated with \texttt{spaCy}.} and the average number of conversations held by each user per day. For Bing Copilot data, we report metrics relative to the first day in the sample or, when comparing activity groups, to users active only one day.

\subsection{Semantic Features}

We also evaluate several high-level semantic properties of each conversation using LLM classifiers: user intent, conversational domain, and task completion. User intent is defined as the user's purpose for conversing with the chatbot and falls into one of nine categories, which include summarization and information lookup. Conversational domain is defined as the general subject area a conversation falls under and is labeled as one of thirty categories, including law and politics and entertainment.  Task completion describes whether the primary task a user intends to accomplish during a conversation is completed by its end (e.g., if the model is prompted to describe photosynthesis, it does so). All three properties are annotated at the conversation level. \Cref{sec:promptAppendix} includes the full text of each classification prompt, including the full list of intent and domain labels and their definitions.

We used GPT-4o-mini with temperature 0 and chain-of-thought prompting~\citep{wei2022chain} for all semantic classification tasks. 
The user intent and conversational domain classification prompts are closely based on those from \citet{wan2024tnt}, who found moderate-to-high agreement between labels produced by GPT-4 and human-annotation consensus (Cohen's $\kappa$ 0.57 and 0.70 respectively). We performed validation of the completion task, finding relatively low agreement between two human annotators ($\kappa=0.25$, 62\% of samples). 
LLM agreement with both human annotators was similarly low ($\kappa=0.29$, 52\%; $\kappa=0.12$, 39\%). Reweighting the stratified annotation sample to account for the highly non-uniform distribution of LLM labels, we estimate that the annotators would agree with the LLM labels on 72\% and 81\% of conversations respectively. From human annotation of the WildChat-4.8M conversations, we found that instances labeled as ``incomplete'' by the LLM were largely driven by cases where the user had no concrete task: e.g., the user said only ``Hi.'' We therefore interpret completion with caution and view it as a proxy for task concreteness. \Cref{app:completion-annotation} contains further details. Our code and classifications for WildChat-4.8M are available at \url{https://github.com/microsoft/adopt-adapt-wildchat}.

\section{Population-Level Trends}

\begin{figure}[t]
    \centering
    \includegraphics[width=0.485\linewidth]{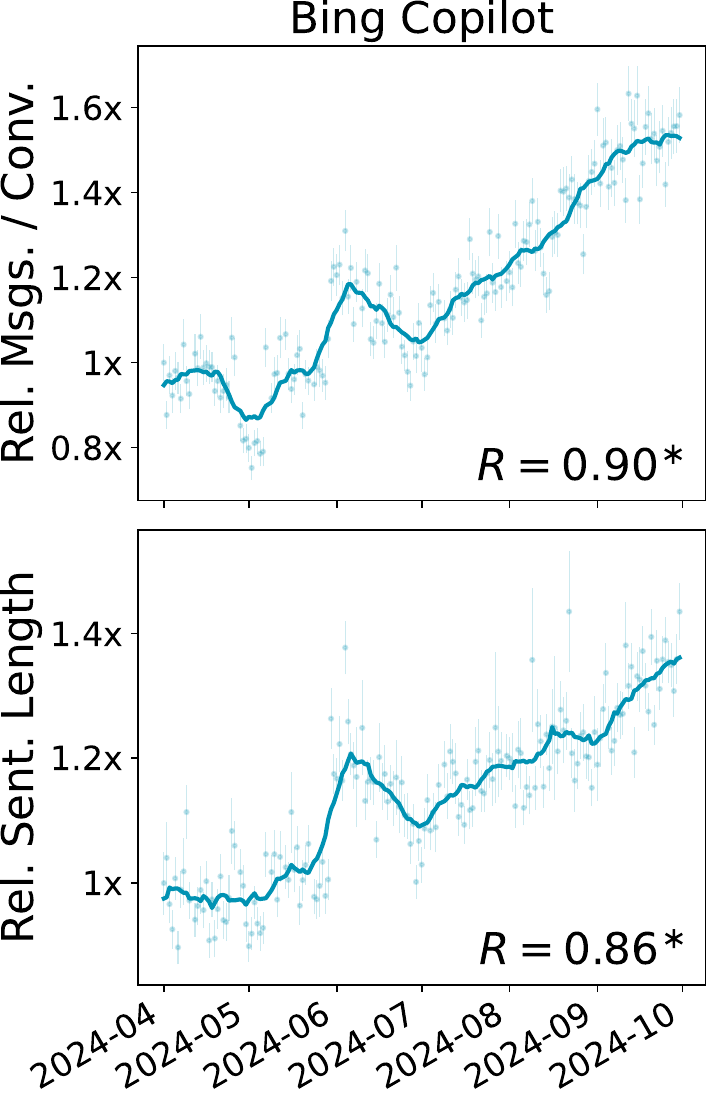}
        \includegraphics[width=0.495\linewidth]{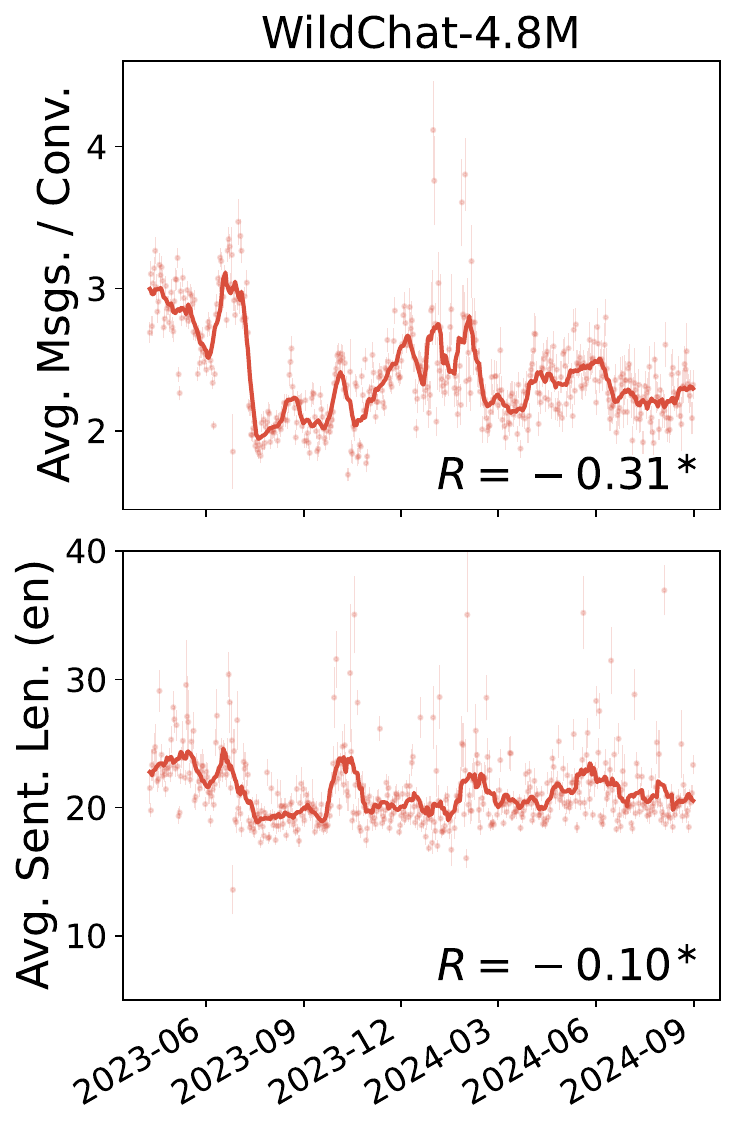}
    \caption{\textbf{Population-level user activity and linguistic complexity rise over time in Bing Copilot, but not WildChat.} Solid lines are 14-day averages; points are daily metrics with standard error. Metrics in Bing Copilot are reported relative to the first day in the sample. Asterisks here and in all other plots indicate Pearson's $R$ is significantly different from 0 ($p < 0.05$).}
    \label{fig:randomConvInfo}
\end{figure}

\begin{figure}[h!]
    \centering
    \includegraphics[width=0.49\linewidth]{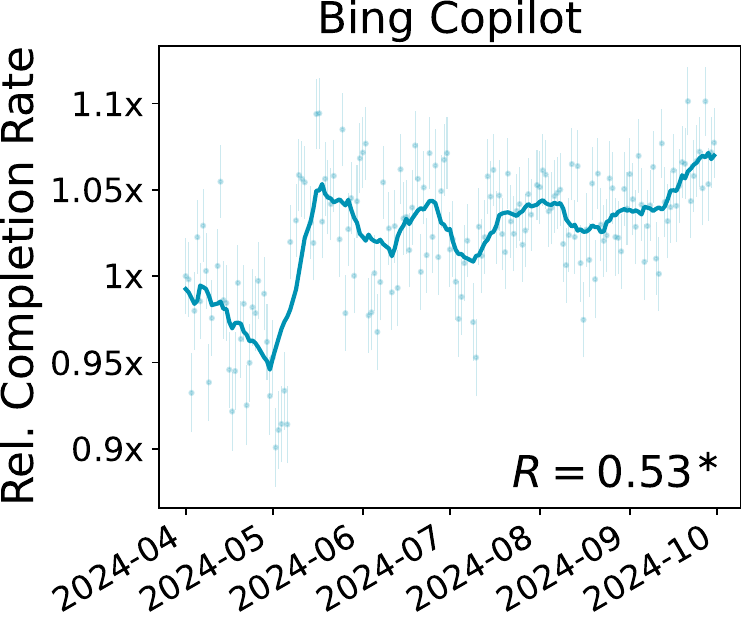}
        \includegraphics[width=0.49\linewidth]{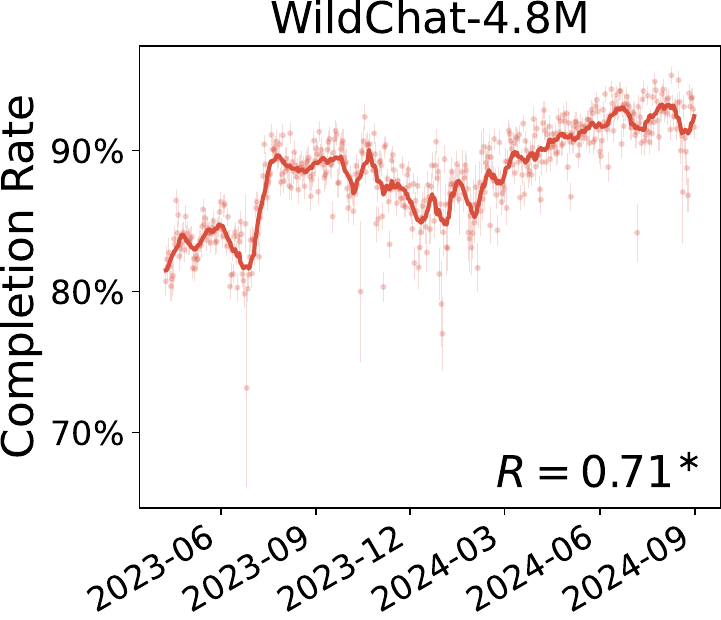}
    \caption{\textbf{Population-level task completion increases over time.}  Solid lines are 14-day averages; points are daily metrics with standard error. For Bing Copilot, completion is reported relative to the first sampled day. }
    \label{fig:randomCompletion}
\end{figure}

\begin{figure}[h!]
    \centering
    \includegraphics[width=0.49\linewidth]{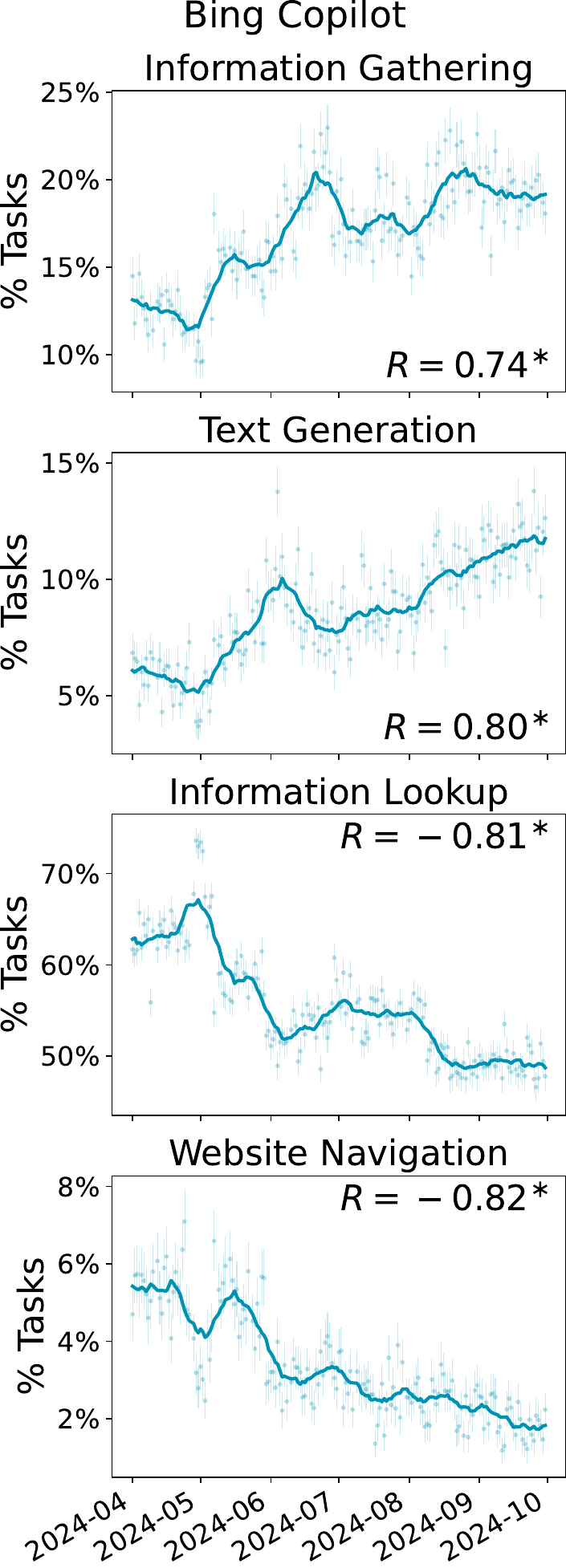}
    \includegraphics[width=0.49\linewidth]{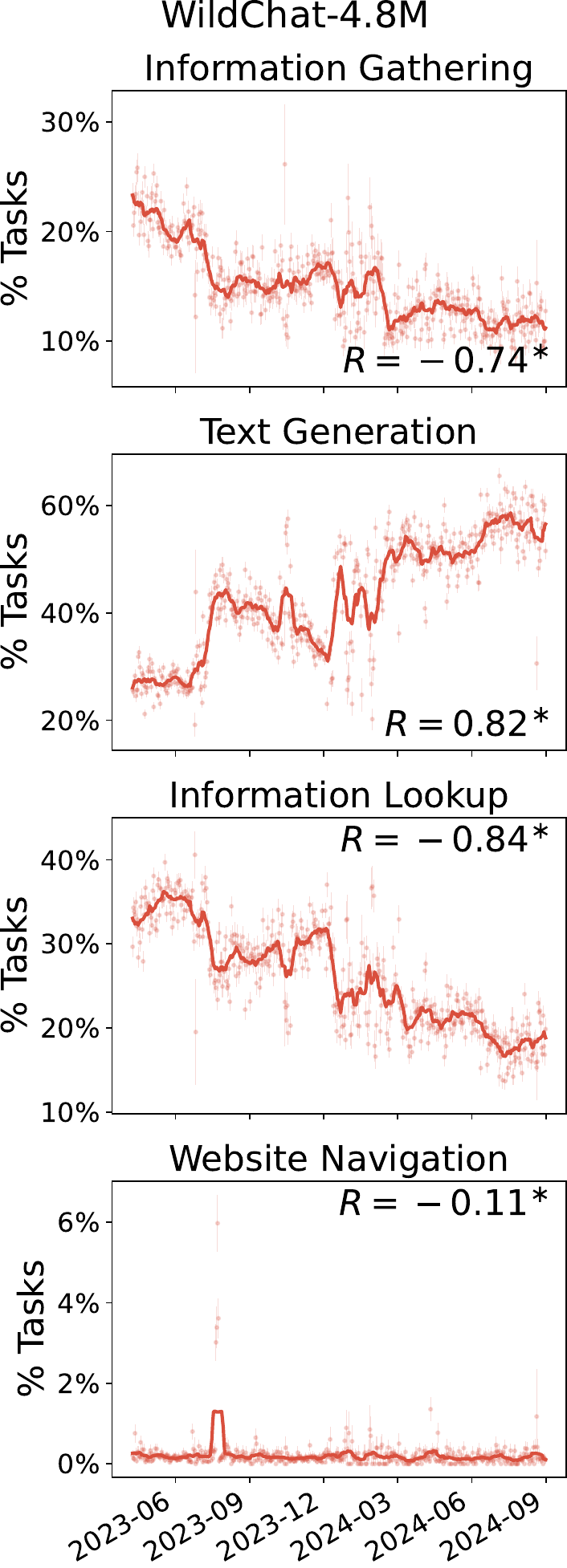}
    \caption{\textbf{For Bing Copilot, more complex tasks rise in population-level popularity over the study period while some simpler tasks fall; these trends only occasionally hold in WildChat.} Solid lines are 14-day averages; points are daily metrics with standard error. Plots of the other intents can be found in \Cref{sec:fig3Appendix}, \Cref{fig:randomIntentsFull}.}
    \label{fig:randomIntents}
\end{figure}

\begin{figure}[h!]
    \centering
    \includegraphics[width=0.51\linewidth]{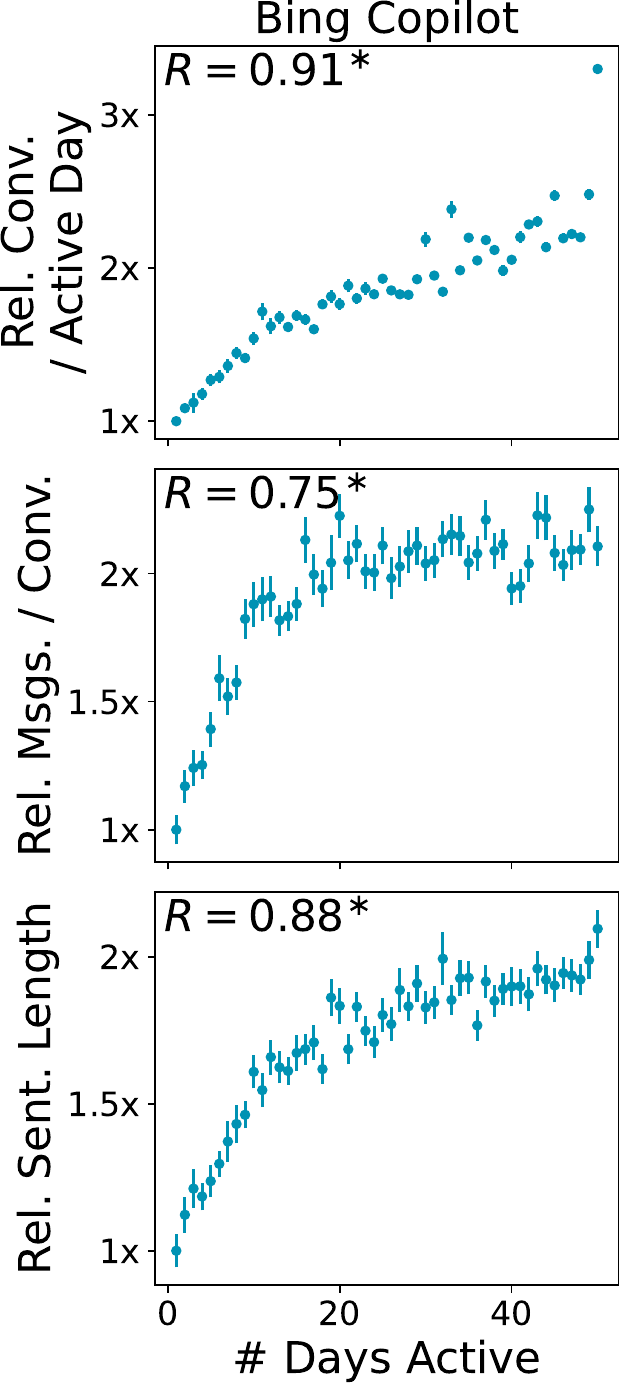}
    \includegraphics[width=0.477\linewidth]{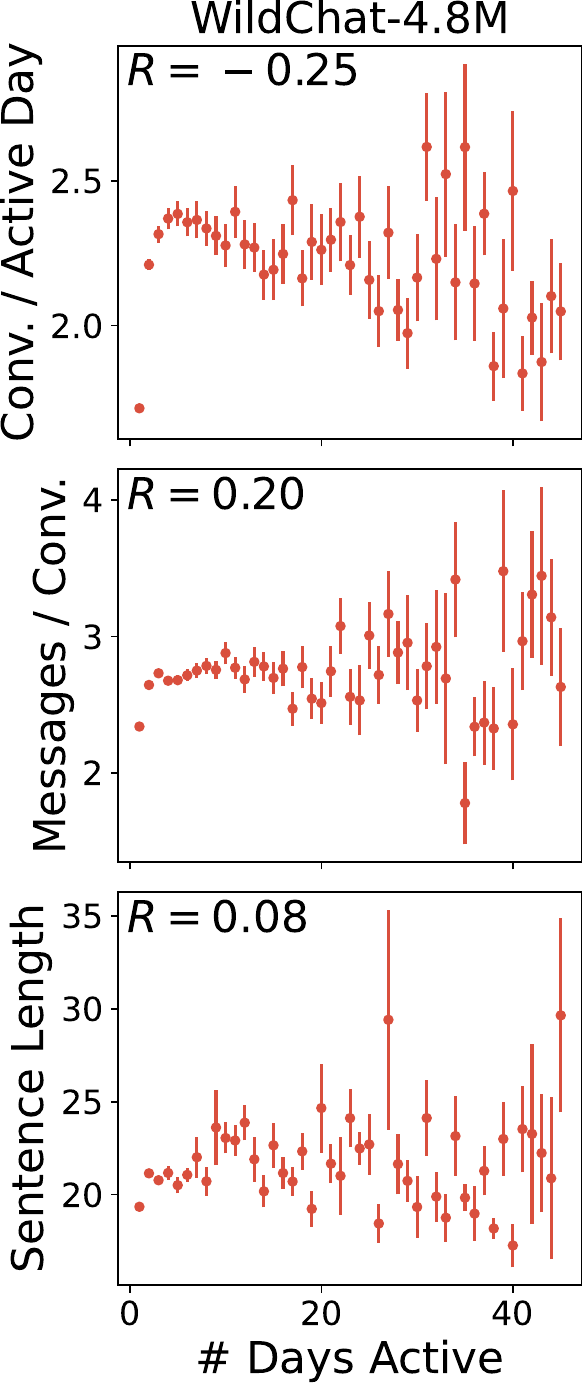}
    \caption{\textbf{More active Bing Copilot users by \# days are more active by two additional metrics and write more linguistically complex messages; the same trends do not hold for Wildchat users.} Bing Copilot metrics are reported relative to users active one day. Error bars represent standard error. WildChat likely shows weaker relationships and more noise at high activity levels due to a lack of stratified sampling.}
    \label{fig:activeDaysConvInfo}
\end{figure}

We first examine population-level trends in the behavior of both user groups during the relevant study periods. In the Bing Copilot population dataset, the average number of user messages per conversation and user message complexity both increase about 1.5$\times$ throughout the study period (\Cref{fig:randomConvInfo}). WildChat does not follow the same trends; average user messages per conversation and average length of user sentences in English both drop slightly. In contrast, task completion increases significantly in both user groups over time (\Cref{fig:randomCompletion}). However, these changes are small and the absolute completion rate is high throughout the WildChat data.

The frequency with which both user populations pursue different task intents also changes over time. Throughout the study period, Bing Copilot users increasingly attempt more complex tasks like information gathering (on average 12.7\% of tasks during the first ten days to 19.2\% during the last ten days) and text generation (6.1\% to 11.6\%) and engage less frequently with simpler tasks like information lookup\footnote{\emph{Information lookup} describes simple factual queries, while \emph{information gathering} describes more complex research.} (62.4\% to 49.2\%) and website navigation (5.3\% to 1.7\%) (\Cref{fig:randomIntents}). Some similar trends are found in the WildChat data: text generation gains popularity (22.8\% to 55.3\%) whereas information lookup loses it (32.2\% to 16.6\%). However, information gathering becomes \textit{less} common in the WildChat data over time (22.7\% to 11.3\%) and website navigation is almost never attempted. Overall, the population-level frequency of intents varies considerably between the datasets.

\begin{figure}[t]
    \centering
    \includegraphics[width=0.498\linewidth]{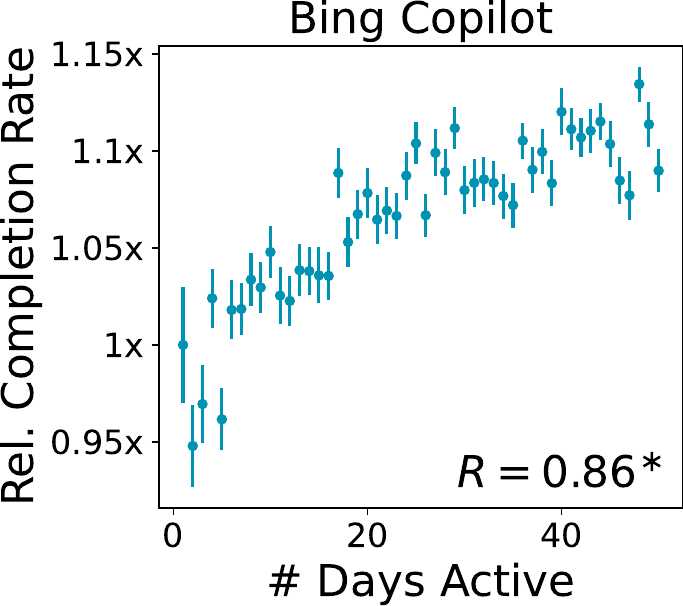}
        \includegraphics[width=0.488\linewidth]{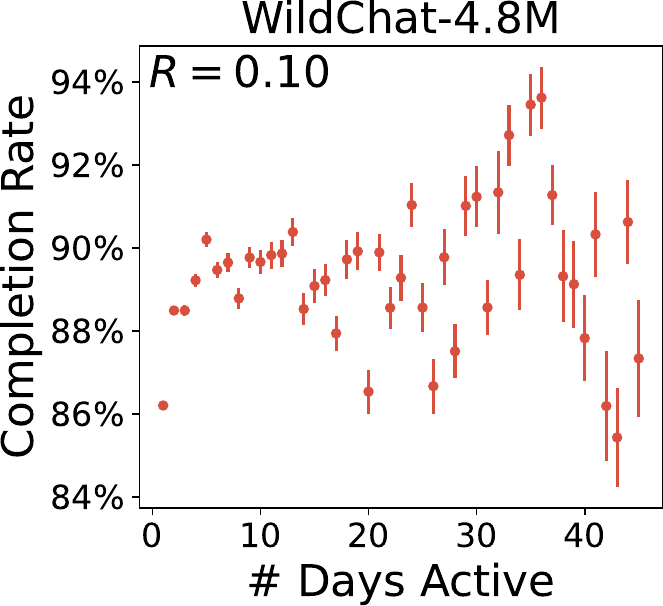}
    \caption{\textbf{More active Bing Copilot users complete more tasks; the same does not hold for WildChat users.} Bing Copilot metrics are reported relative to users active one day. Error bars represent standard error.}
    \label{fig:activeDaysCompleted}
\end{figure}

\section{Differences by Activity Level}

Next, we examine whether differences exist between users of varying activity levels. To do this, we first average feature values for each user on every day they are active. Then, we average over all days a user is active. Finally, we average all users in a single activity level group (defined by the number of days the user is active). Thus, each day a user is active is weighted equally, although the weight of individual conversations may vary based on a user's activity level per day.

We find clear differences between the behavior of Bing Copilot users of different activity levels, but not between WildChat users. Bing Copilot users who are active on more days also hold more conversations per day, send more messages per conversation, and write more linguistically complex messages (\Cref{fig:activeDaysConvInfo}). None of these trends are found in the WildChat data, although a jump in all three metrics occurs between users who are active on only one day and those who are active on $\geq$2 days. The increased noise found in high-activity WildChat data is likely due to a lack of stratified sampling and the resulting small sample sizes for high active day groups; similarities between users of different activity levels may be due in part to the imprecise method of identifying individual users by hashed IP.

\begin{figure}[h]
    \centering
    \includegraphics[width=\linewidth]{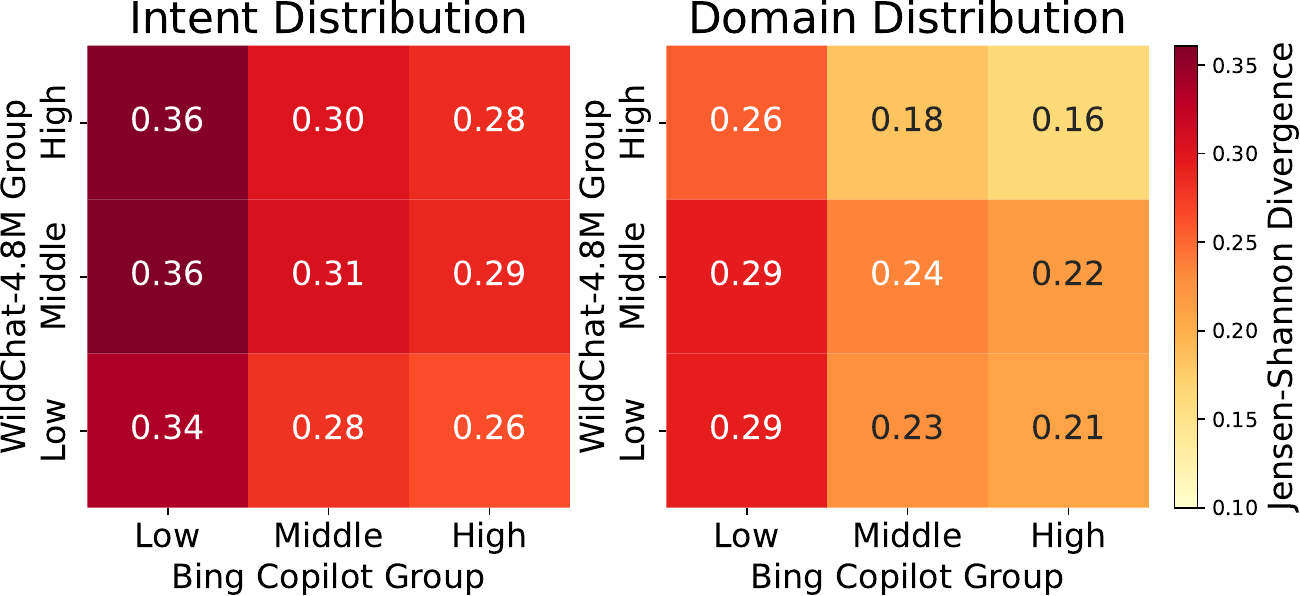}
    \caption{\textbf{WildChat usage more closely resembles high-activity Bing Copilot users.} The heatmaps show the Jensen--Shannon divergence (smaller = more similar) between the intent (left) and domain (right) distributions of Bing Copilot and WildChat users in each activity group.}
    \label{fig:dsn-heatmaps}
\end{figure}

\begin{table*}[t]
    \small
    \centering
\definecolor{posblue}{HTML}{4393C3}
\definecolor{negorange}{HTML}{D6604D}
\begin{tabular}{lccccc}
    \toprule
    & \multicolumn{2}{c}{\textbf{Bing Copilot}} & \multicolumn{3}{c}{\textbf{WildChat-4.8M}} \\
    \cmidrule(lr){2-3} \cmidrule(lr){4-6}
    \textbf{Intent} & \textbf{Middle} (pp) & \textbf{High} (pp) & \textbf{Low} (\%) & \textbf{Middle} (pp) & \textbf{High} (pp) \\
    \midrule
    Summarization & \cellcolor{posblue!55}$+15.7$ & \cellcolor{posblue!68}$\bm{+19.3}$ & $91.8$ & \cellcolor{posblue!4}$+1.2$ & \cellcolor{posblue!11}$\bm{+3.2}$ \\
    Translation or Conversion & \cellcolor{posblue!27}$+7.7$ & \cellcolor{posblue!53}$\bm{+15.1}$ & $89.8$ & \cellcolor{posblue!10}$\bm{+2.9}$ & \cellcolor{posblue!11}$\bm{+3.0}$ \\
    Open-Ended Discovery & \cellcolor{posblue!22}$+6.2$ & \cellcolor{posblue!45}$\bm{+12.8}$ & $70.0$ & \cellcolor{negorange!10}$-2.8$ & $+0.0$ \\
    Analysis & \cellcolor{posblue!10}$+2.8$ & \cellcolor{posblue!27}$\bm{+7.9}$ & $86.2$ & \cellcolor{negorange!1}$-0.3$ & \cellcolor{posblue!3}$+0.9$ \\
    Information Gathering & \cellcolor{posblue!21}$\bm{+5.9}$ & \cellcolor{posblue!26}$\bm{+7.5}$ & $86.1$ & \cellcolor{posblue!1}$\bm{+0.4}$ & \cellcolor{posblue!6}$+1.7$ \\
    Text Generation & \cellcolor{posblue!7}$+2.1$ & \cellcolor{posblue!16}$\bm{+4.6}$ & $94.4$ & \cellcolor{negorange!6}$\bm{-1.8}$ & \cellcolor{negorange!8}$\bm{-2.2}$ \\
    Information Lookup & \cellcolor{posblue!9}$\bm{+2.4}$ & \cellcolor{posblue!13}$\bm{+3.6}$ & $85.8$ & \cellcolor{posblue!4}$\bm{+1.2}$ & \cellcolor{posblue!5}$\bm{+1.4}$ \\
    Image Creation & $+0.0$ & \cellcolor{posblue!1}$\bm{+0.3}$ & $83.4$ & \cellcolor{posblue!13}$\bm{+3.6}$ & \cellcolor{posblue!2}$\bm{+0.5}$ \\
    Website Navigation & \cellcolor{negorange!4}$-1.2$ & \cellcolor{negorange!17}$-4.8$ & $62.4$ & \cellcolor{posblue!23}$\bm{+6.6}$ & \cellcolor{posblue!53}$+15.2$ \\
    \bottomrule
\end{tabular}
    \caption{\textbf{More active users have higher completion rates for most intents, particularly in Bing Copilot.} The table shows the percentage point (pp) difference in completion rates per intent between middle/high and low activity groups. Baseline low-activity completion rates are also reported for WildChat. Bold numbers indicate a significant two-proportion $Z$-test vs the low activity group ($p < 0.05$ after Bonferroni correction within each dataset).}
    \label{tab:activeDaysIntentCompCorrs}
\end{table*}

\begin{table*}[t]
    \small
    \centering
\definecolor{posblue}{HTML}{4393C3}
\definecolor{negorange}{HTML}{D6604D}
\begin{tabular}{lcccccc}
    \toprule
    & \multicolumn{3}{c}{\textbf{Bing Copilot}} & \multicolumn{3}{c}{\textbf{WildChat-4.8M}} \\
    \cmidrule(lr){2-4} \cmidrule(lr){5-7}
    \textbf{Domain} & \textbf{Low} (\%) & \textbf{Middle} (pp) & \textbf{High} (pp) & \textbf{Low} (\%) & \textbf{Middle} (pp) & \textbf{High} (pp) \\
    \midrule
    Programming and Scripting & $5.1$ & \cellcolor{posblue!36}$\bm{+2.6}$ & \cellcolor{posblue!54}$\bm{+3.8}$ & $14.5$ & \cellcolor{posblue!14}$\bm{+1.0}$ & \cellcolor{negorange!19}$\bm{-1.4}$ \\
    Creative Writing and Editing & $3.5$ & \cellcolor{posblue!28}$\bm{+2.0}$ & \cellcolor{posblue!40}$\bm{+2.9}$ & $21.6$ & \cellcolor{negorange!36}$\bm{-2.6}$ & \cellcolor{negorange!70}$\bm{-6.7}$ \\
    Professional Writing and Editing & $1.0$ & \cellcolor{posblue!10}$\bm{+0.7}$ & \cellcolor{posblue!20}$\bm{+1.4}$ & $1.0$ & \cellcolor{posblue!4}$\bm{+0.3}$ & \cellcolor{posblue!24}$\bm{+1.7}$ \\
    Entertainment & $4.8$ & \cellcolor{negorange!22}$\bm{-1.6}$ & \cellcolor{negorange!10}$-0.7$ & $2.8$ & \cellcolor{posblue!6}$\bm{+0.5}$ & \cellcolor{negorange!8}$-0.6$ \\
    Shopping and eCommerce & $1.7$ & \cellcolor{negorange!8}$\bm{-0.6}$ & \cellcolor{negorange!12}$\bm{-0.8}$ & $0.6$ & \cellcolor{posblue!7}$\bm{+0.5}$ & \cellcolor{posblue!10}$\bm{+0.7}$ \\
    Small Talk and Chatbot & $2.4$ & \cellcolor{negorange!6}$\bm{-0.4}$ & \cellcolor{negorange!12}$\bm{-0.9}$ & $2.4$ & \cellcolor{posblue!5}$\bm{+0.4}$ & \cellcolor{negorange!2}$\bm{-0.2}$ \\
    Travel and Tourism & $3.2$ & \cellcolor{negorange!24}$\bm{-1.7}$ & \cellcolor{negorange!28}$\bm{-2.0}$ & $0.6$ & $\bm{+0.0}$ & \cellcolor{posblue!1}$+0.0$ \\
    \bottomrule
\end{tabular}
    \caption{\textbf{More active Bing Copilot users are more likely to converse about professional topics and less likely to converse about casual topics; professional topics are more popular for all WildChat activity levels.} The table shows the difference in conversation share belonging to a subset of domains between middle/high and low activity users, measured in percentage points (pp) over low \%. Bold numbers indicate a significant two-proportion $Z$-test vs the low activity group ($p < 0.05$ after Bonferroni correction within each dataset). An expanded table with all domains can be found in \Cref{sec:domainActiveAppendix}, \Cref{tab:activeDaysDomainsFull}.}
    \label{tab:activeDaysDomains}
\end{table*}

More active Bing Copilot users also appear to complete more tasks (\Cref{fig:activeDaysCompleted}). This pattern largely holds even when broken down by task intent; we find that high activity Bing Copilot users have significantly more completed tasks than low activity users of all intents types but website navigation (\Cref{tab:activeDaysIntentCompCorrs}). While the differences are rarely significant, the same pattern often holds for middle activity users. Interestingly, the tasks for which the differences in completion rates between users of different activity levels are largest are often more complex. More active users may frame these complex tasks more \emph{concretely}, which we found is often conflated with completion by the annotating LLM.

In contrast, task completion only increases with activity level for WildChat users active $\leq$5 days (\Cref{fig:activeDaysCompleted}). For some intents, completion increases with user activity, but these changes are generally small and there is no clear association with task complexity (\Cref{tab:activeDaysIntentCompCorrs}). In fact, website navigation, the only intent for which higher activity Bing Copilot users are not more successful, is the intent for which this trend is strongest in WildChat users. For all other intents, $\geq$70\% of tasks are completed even for low activity users.

Additionally, we find that high activity Bing Copilot users are significantly more likely to have conversations that fall into professional or creative domains (e.g., professional writing \& editing, programming \& scripting, creative writing \& editing) and are less likely to engage in more entertainment-oriented tasks (e.g., entertainment, travel \& tourism, shopping \& eCommerce, small talk \& chatbot) (\Cref{tab:activeDaysDomains}). In WildChat, even low-activity users have high rates of programming and creative writing and lower rates of shopping, travel, and entertainment. WildChat users overall most closely resemble \textit{high} activity Bing Copilot users. We confirm this by examining the Jensen--Shannon (J--S) divergence between the task intent and conversational domain distributions of low, middle, and high activity WildChat and Bing Copilot users (\Cref{fig:dsn-heatmaps}). Overall, the J--S divergences between the WildChat intent and domain distributions and low-activity Bing Copilot users are 0.34 and 0.28 respectively, but divergences are only 0.26 and 0.19 with high-activity Bing Copilot users.

\section{Changes Over User Trajectories}

\begin{figure*}[h!]
    \centering
    \includegraphics[width=.458\linewidth]{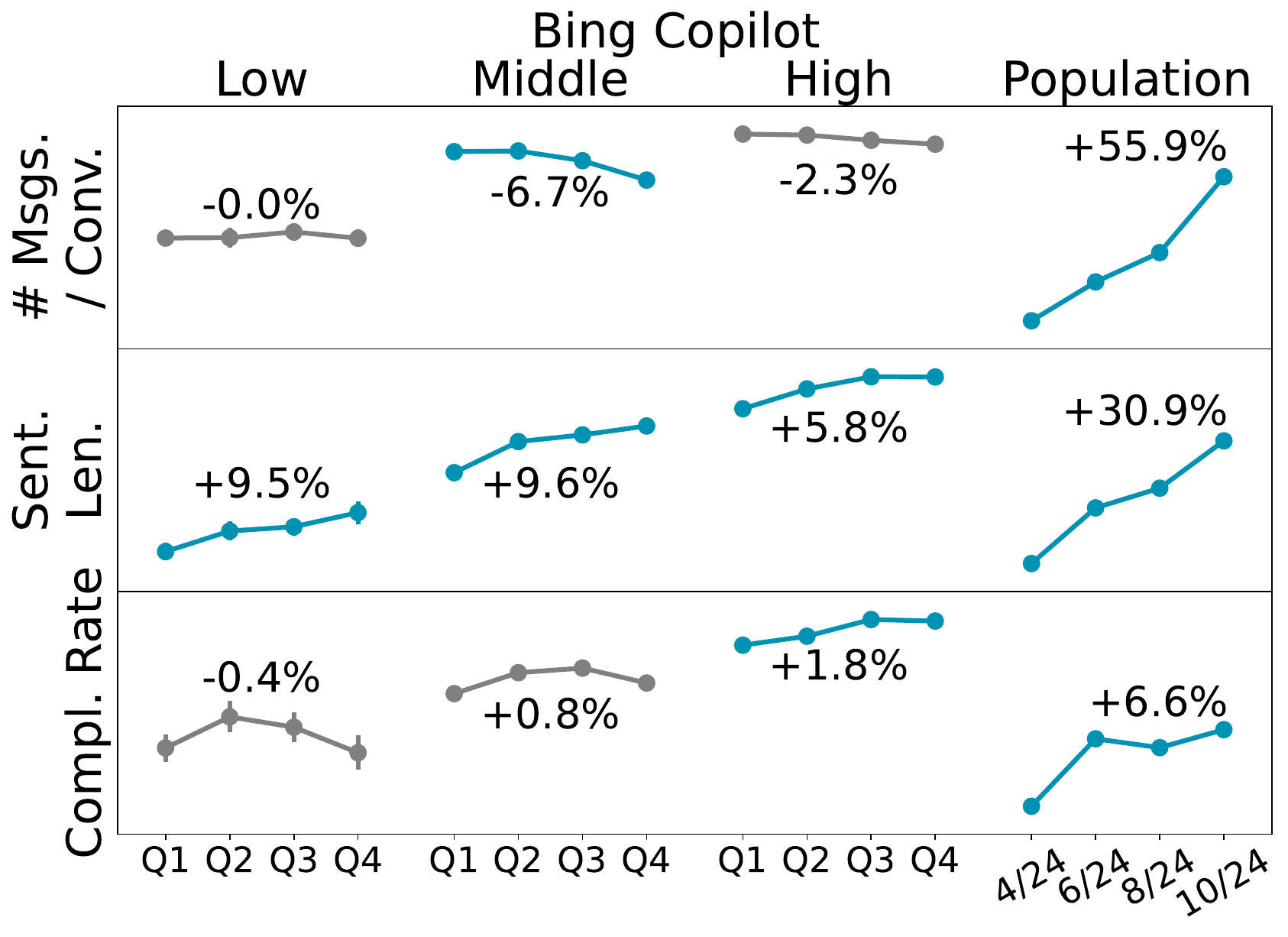}
    \includegraphics[width=0.52\linewidth]{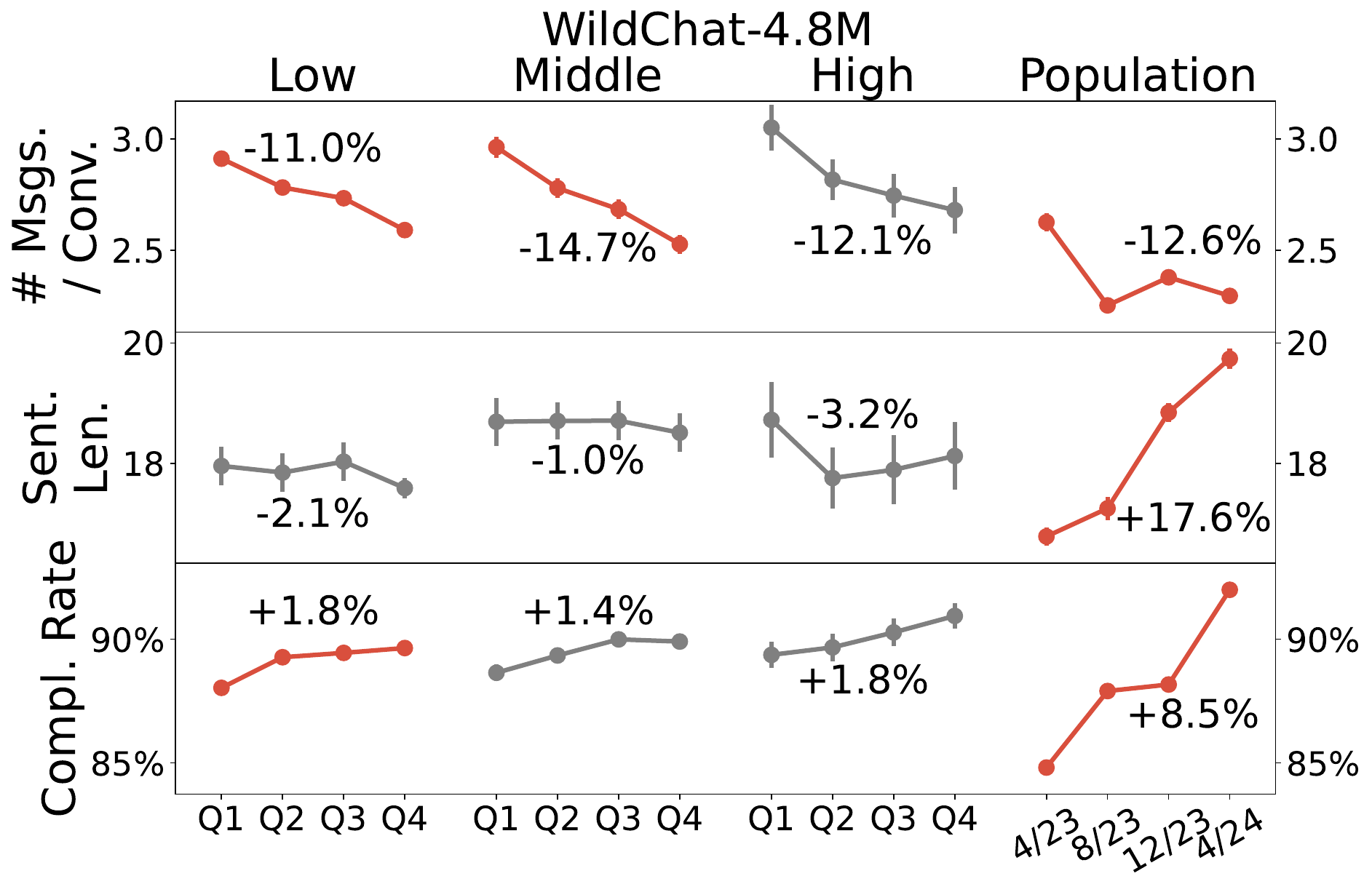}    \caption{\textbf{Over user lifetimes, changes in activity, linguistic complexity, and completion are usually smaller than at the population level.} Average feature values during each quarter of user trajectories (e.g., Q1 = first quarter of days active), stratified by activity level. Users active for fewer than four days are dropped so quarters are meaningful. Population values are plotted temporally. Error bars represent standard error. Markers are colored rather than gray if a paired $t$-test for difference in means between the first and last quarter is significant ($p < 0.05$) after Bonferroni correction within each dataset.}
    \label{fig:activityLevelLifetimeChanges}
\end{figure*}

Having established that significant differences exist between Bing Copilot users of different activity levels, we now probe to see whether these differences are inherent (present from the beginning of a user's conversational trajectory) or learned (developed over the user's trajectory). The strong population-level trends we see in \Cref{fig:randomConvInfo,fig:randomCompletion,fig:randomIntents} could suggest that individual users are evolving in their LLM usage over time. Surprisingly, we find that this is not the case, and that population-level trends are caused by the arrival of new users with different habits. 

To study user trajectories, we first index every day each user is active; the first day they appear in the dataset is labeled 0, then 1, and so on. We filter out all days with an index higher than 49 for the Bing Copilot data and 44 for the WildChat data. Then, we divide each user trajectory into quarters and find the mean feature values for each quarter (labeled Q1--Q4). Next we split all the users into activity-level classes: users active 1--10 days (low), users active 11--25 days (middle), and users active 26+ days (high). For each activity-level class, we run a paired $t$-test for difference in means comparing the mean feature values from the first and last quarters of the trajectory of each user. For comparison, we also visualize population-level trends, splitting the study period in four and plotting the average feature values from the population dataset for each quarter of the study period.

Perhaps surprisingly, we find that most changes over user trajectories are quite small, though some are significant. The average number of messages per conversation for middle activity Bing Copilot users and low--middle activity WildChat users decreases significantly over time (\Cref{fig:activityLevelLifetimeChanges}). This decrease is mirrored at the population level in WildChat, but is the opposite of the Bing Copilot population trend.

In contrast, while the population's linguistic complexity increases significantly over time for both datasets, this trend appears on the user-level only in Bing Copilot and, even then, the increase for each activity group is much smaller than that of the population.
Similarly, while completion increases over time in both datasets, these changes are smaller in user trajectories than in the population  and not always significant (\Cref{fig:activityLevelLifetimeChanges}). 

We find evidence that the explanation for these apparent contradictions (relatively flat within-user trajectories but substantial population-level trends) is that users who start using Bing Copilot later behave substantially differently than users who join earlier. In \Cref{fig:userBehaviorByStartDate} (\Cref{app:start-date}), we show the same metrics as in \Cref{fig:activityLevelLifetimeChanges}, but only for users active 1--4 days (who represent a substantial distribution of the population due to the power law distribution of usage), grouped by the temporal quarter of the data in which their first conversation occurs. Users who join the platform later reflect the population-level trends in \Cref{fig:activityLevelLifetimeChanges}, highlighting that these trends are driven by new users with different habits rather than within-user changes.

There are also few consistent changes in the task intents users pursue throughout their conversational trajectories (\Cref{sec:intentTrajecAppendix}, \Cref{fig:intentDayCorrsQuart}). In the Bing Copilot data, some significant changes in intent pursuit align with population-level trends: middle and high activity users decrease in proportion of website navigation tasks and high activity users increase in pursuit of text generation. However, some user-level changes reverse population-level patterns: middle and high activity users decrease their pursuit of image creation and high activity users decrease their proportion of open-ended discovery.
Notably, no significant changes in intent pursuit occur over the lifetimes of low activity users and even the significant changes in higher activity user behavior are usually much smaller in magnitude than the population level changes (again, with population-level effects driven by the arrival of new users; \Cref{fig:userIntentsByStartDate}, \Cref{app:start-date}).

\begin{figure*}[t]
    \centering
    \includegraphics[width=.48\linewidth]{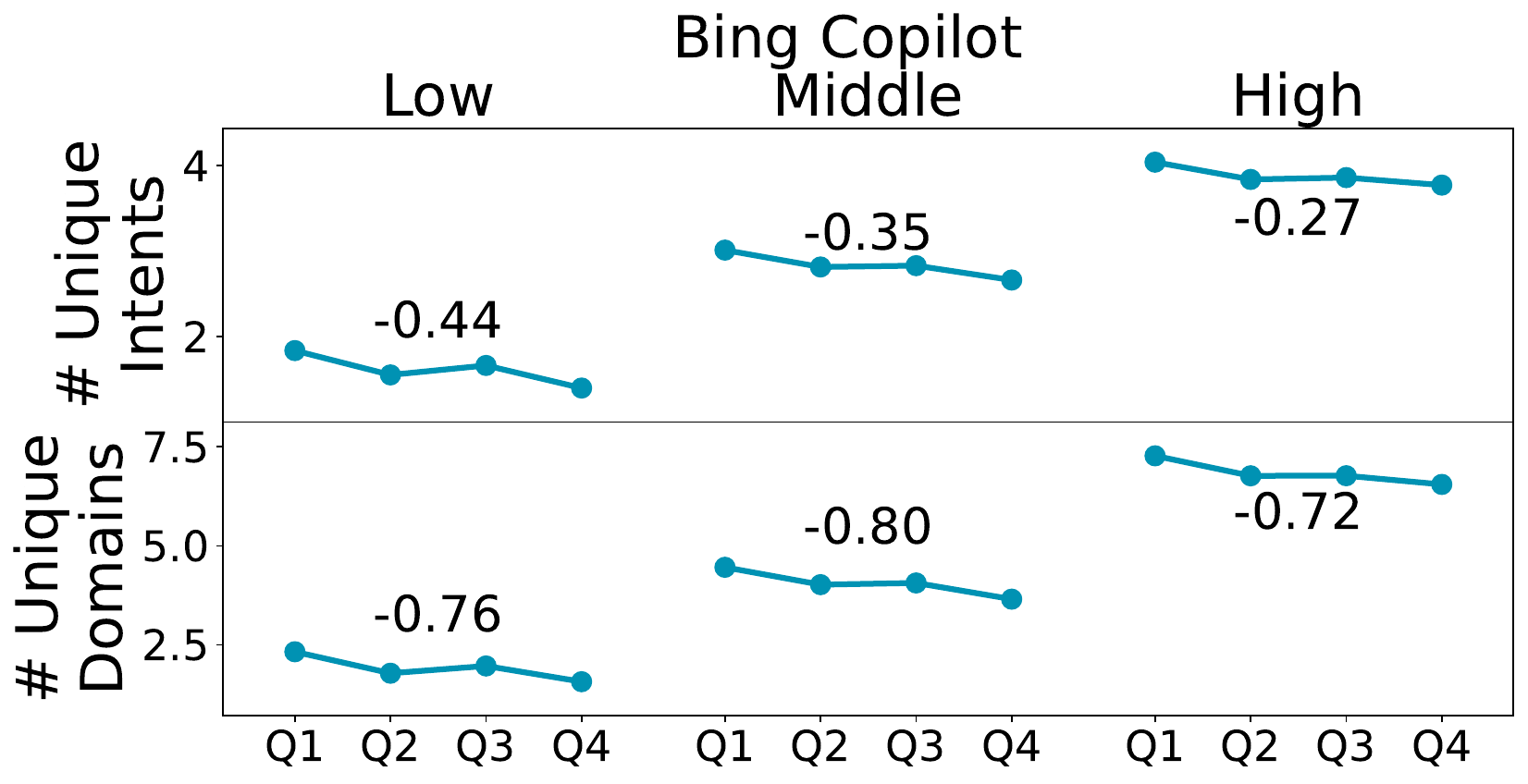}%
    \hfill
    \includegraphics[width=.49\linewidth]{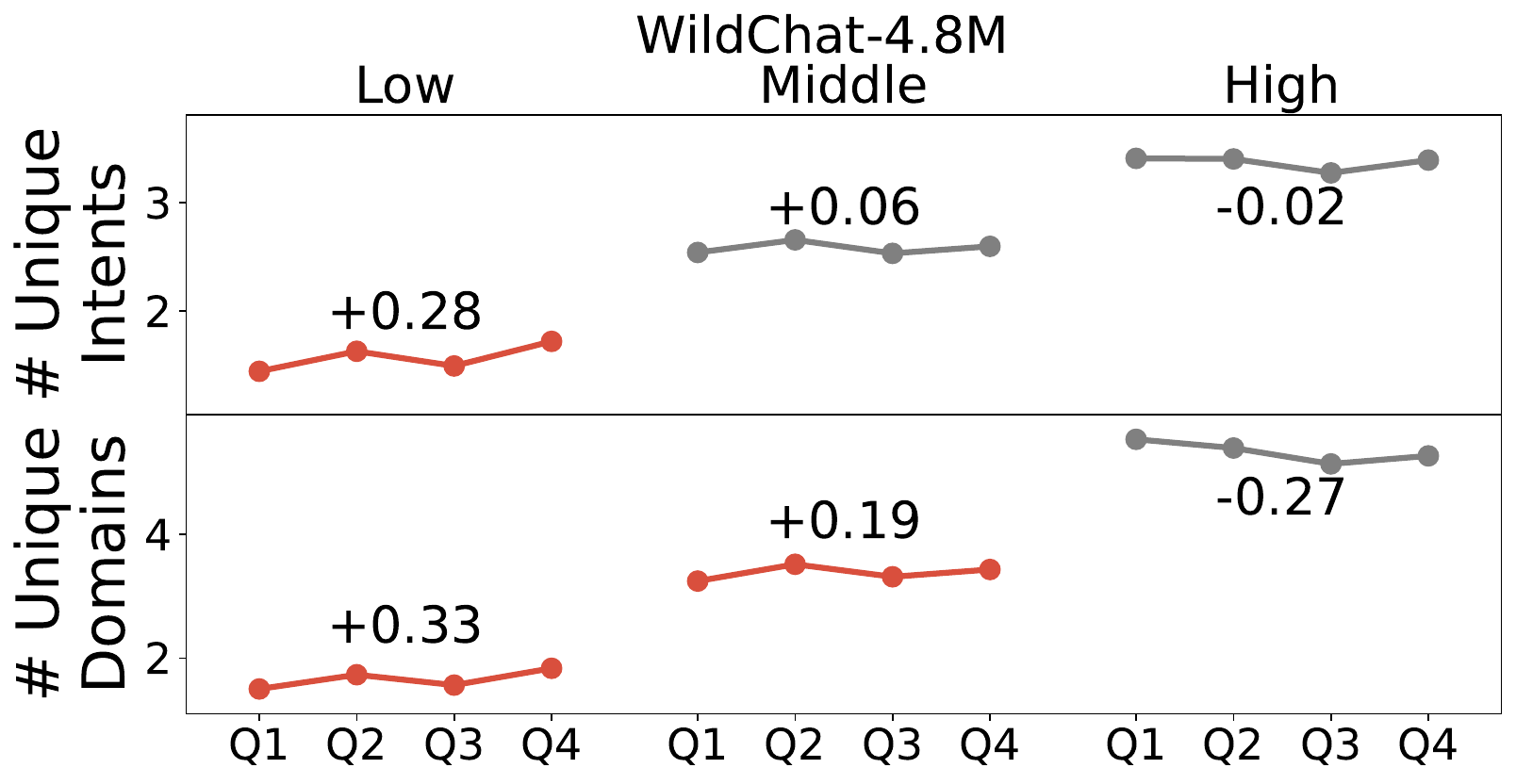}
    \caption{\textbf{Bing Copilot users shift from exploration to exploitation over their lifetimes; lower activity WildChat users do the reverse.} Average \# unique intents (left) and domains (right) during each quarter of user trajectories. Error bars represent standard error (smaller than the markers). Markers are colored rather than gray if a paired $t$-test for difference in means between the first and last quarter is significant ($p < 0.05$) after Bonferroni correction within each dataset.}
    \label{fig:numUniqueLifetimes}
\end{figure*}

The significant user-level changes in intent pursuit found in the Bing Copilot data only sporadically align with changes in WildChat.
Again, there is a significant increase in the pursuit of text generation, but only among low and middle activity users, and there is a significant decrease in the proportion of open-ended discovery tasks by low and high activity users. However, while there is no significant change in information gathering for Bing Copilot users, its pursuit decreases for low and middle activity WildChat users.
Again, the changes over WildChat user trajectories are usually much smaller in magnitude than the population level changes. In both datasets, we find that users change their behavior comparatively little over time, and what changes do exist do not align closely enough over the datasets to suggest general trends.

As with intent, there are few significant changes in the conversational domains users from both datasets pursue throughout their interactions with the chatbots (\Cref{sec:domainTrajecAppendix}, \Cref{fig:domainDayCorrsQuart1,fig:domainDayCorrsQuart2}). Again, these changes are usually small and rarely align across the two datasets. The only consistent pattern is that high activity Bing Copilot users and low and high activity WildChat users both significantly decrease small talk over time, although the change is very small as the domain is rare.

Finally, we explore whether users shift between exploration and exploitation of LLM uses over their trajectories. We measure this by counting the number of unique intents and domains that each user pursues in the each quarter of their trajectories. Then, we separate users by activity-level class and apply paired $t$-tests for difference in means to compare the number of unique intents and domains explored in  their first and last quarters.

For all activity-level classes of Bing Copilot users, we find there is a significant $\sim$10--30\% decrease in the number of unique intents and domains pursued over time (\Cref{fig:numUniqueLifetimes}); these users show a shift from exploration to exploitation, paralleling the findings of \citet{zhu-etal-2026-show,zhu2026priming} on crystallization of habits. However, in WildChat, we find that low and middle activity users explore more over time while high activity users have no significant change.

\section{Conclusion}

There are many enticing hypotheses for how LLM user behavior may change over time. However, in practice, we find only small changes in five dimensions of user behavior throughout users' interactions with Bing Copilot and the WildChat interface.
Our findings further reinforce the centrality of habits highlighted by \citet{zhu2026priming}, who provide an excellent discussion of the causes and implications of the stickiness of LLM user habits.

The changes that we do find over Bing Copilot user trajectories are much smaller than the differences between less and more active users. From their first conversations with Bing Copilot, more active users complete more tasks, pursue different intents, and write more linguistically complex prompts, suggesting that differences between users are more inherent than learned. In the future, in order to help users adapt to changing technology and discover its most useful applications, it will be important to proactively help them break out of habits.

We also find that Bing Copilot users differ from those in the WildChat dataset in many salient ways. WildChat users most closely resemble the most active Bing Copilot users, but differ even from this group in how their behavior changes over time. Additionally, the WildChat dataset contains a large fraction of API-like non-conversational usage. Some of the changes observed between the datasets may be a result of differences in the way each dataset was processed; nonetheless, the differences are dramatic enough that we encourage careful consideration of WildChat's composition before using it in downstream applications, particularly when trying to mimic average LLM user behavior.

\section*{Limitations}

We wish to highlight several limitations of this work. First, this study only includes data from Bing Copilot taken between January 1 and September 30, 2024 and from WildChat-4.8M (focusing on data before September 2024 due to spikes in API-like usage later in the data). This usage may differ significantly from that on different AI platforms during different periods. In addition, our measures of success and complexity do not capture the full spectrum of user satisfaction and task sophistication. Our analysis of user trajectories focuses on patterns shared across users (e.g., whether individuals, on average, do more of a certain intent as they use the LLM more); it is possible that different users each evolve in different directions, in ways that do not show up in overall averages. Finally, our dataset has an arbitrary end date, meaning we do not capture the entirety of all user trajectories. We have no definitive way of knowing when a user's \textit{last} interaction with the chatbot is; some changes in user behavior may occur after our dataset ends.

In the WildChat data, our usage of hashed IP addresses to link conversations that may belong to the same user is imperfect, likely missing conversations from the same person when devices or networks change and possibly conflating conversations from different people sharing an IP address in ways that our filter did not exclude. It is possible that the weaker distinction between different activity level groups in WildChat is in part driven by the weaker user--conversation mapping than we have in the Bing Copilot data. We also chose not to remove ``templated'' API-like conversation from WildChat, opting only to use the date cutoff to remove the most-affected time period. As a result, some fraction of the conversations in our WildChat analysis are API-like, although plots in \Cref{app:full-wildchat} show that removing templated conversations does not have a significant effect on the WildChat trends we observe.

\section*{Acknowledgments}

This study was approved by Microsoft Research Privacy, Ethics, and Compliance review (\#8986).  We thank Siddharth Suri, Nirupama Chandrasekaran, Jennifer Neville, Axel Bax, and the MSR AI Interaction and Learning group for helpful discussions and feedback. We thank the anonymous reviewers for their useful suggestions, in particular the idea of splitting user cohorts by start date as in \Cref{app:start-date}. We acknowledge the use of AI coding assistants for data analysis and visualization; we take full responsibility for the results and all contents of the paper. 

\bibliography{custom}

\clearpage
\appendix

\section{Prompts}
\label{sec:promptAppendix}

In the following prompts, \texttt{\{content\}} was replaced by the conversation text. User messages were demarcated by \verb+<| start user message |>+ and \verb+<| end user message |>+, while AI messages were demarcated by \verb+<| start agent message |>+ and \verb+<| end agent message |>+. Messages longer than \verb+max_len = 5000+ characters were abbreviated as follows:

\begin{lstlisting}[language=python,showstringspaces=false]
def abbreviate_turn(content, max_len):
    if len(content) < max_len:
        return content

    return content[:max_len//2] + ' [... more text ...] ' + content[-max_len//2:]
\end{lstlisting}

Note that 2500 characters is approximately 400 English words (about 1 page of text), enough to make the intent and context of the messages clear.
\subsection{Completion Prompt}

\begin{lstlisting}
# Task Overview
You will be given a conversation 
between a user and an AI chatbot. You 
will summarize the main task that the 
user is trying to accomplish in the 
conversation. You will also determine 
whether the AI chatbot is able to 
complete the task and whether any 
explicit positive or negative feedback 
is given by the user during the 
conversation.


# Task details
Your task is to fill out the following 
fields:

user_task_summary: A one sentence 
summary of the main task the user is 
trying to accomplish in this 
conversation. Include only the task the 
**user** intends to complete, and not 
the one the agent performs.

completed_explanation: Based on the 
provided conversation, explain in one 
sentence whether the AI chatbot 
completed the user's task.

completed: Based on your explanation, 
provide one of the following labels:
- not_completed: The AI chatbot did not make substantive progress towards completing the user's task.
- partially_completed: The AI chatbot made progress towards completing the user's task, but did not complete it.
- completed: The AI chatbot completed the user's task.

# Hints
Provide your answers in **English** 
using the given structured output 
format.
<| end instructions |>

<| start conversation |>
{content}
<| end conversation |>

<| end prompt |>
\end{lstlisting}

\subsection{Task Intent Prompt}

\begin{lstlisting}
<| start instructions |>
# Task Overview
You will be given a conversation 
between a user and an AI chatbot. Your 
job is to classify the user intent in 
the conversation. User intent is 
defined as the user's purpose for 
conversing with the AI agent. Note that 
the intent is to be based on the user's 
query and not necessarily on the AI 
agent's response.

# Task details
Your task is to fill out the following 
fields:

user_intent_explanation: Based on the 
provided conversation, explain in one 
sentence what the user intent is.

user_intent: Based on your explanation, 
provide one of the following labels:
- web_site_navigation: Conversations where the user simply wants to go to a particular website or service. For example, "facebook" indicates the user wants to go to the Facebook page to logon, "gmail" indicates the user wants to go to Gmail to perform email related actions and "google translate" indicates the user wants to go to "Google Translate" page.
- information_lookup: Conversations where the user wants to find factual information or answers to specific questions. The agent's responses are typically direct, concise, and informative, providing the relevant information and/or links to the sources.
- information_gathering: Conversations where the user wants to gather information that they can use to do their task. This could be asking the agent for recommendations, asking the agent to compare two or more objects, events, ideas, problems, or situations, etc.
- translation_or_conversion: Conversations where the user wants to convert information from one form or representation to another, such as from one language to another, from one unit to another, from words to numbers or vice versa.
- summarization: Conversations where the user wants to get a brief statement that captures the main idea or theme of the given information.
- analysis: Conversations where the user asks analytical or conceptual questions about a complex topic or problem. The user's questions require some degree of reasoning, interpretation, argumentation, comparison, and/or data processing from the agent. The agent's responses are typically explanatory and detailed, and are based on evidence, logic, and facts.
- image_creation: Conversations where the user asks the agent to either generate or modify an image based on specified criteria or constraints.
- text_generation: Conversations where the user asks the agent to either generate new content or modify existing content based on specified criteria or constraints. In the case of generating new content, the user's questions require some degree of creativity, novelty, or innovation from the agent. The agent's responses contain new or modified outputs that match the user's specifications.
- open_ended_discovery: Conversations where the user wants to chat or play with the agent out of curiosity, boredom, or humor, or else explore broad ideas or areas of interest without a specific goal or information need in mind. The agent's responses are typically suggestive and engaging. The agent may also encourage further inquiry or action from the user to deepen their discovery experience.

# Hints
Provide your answers in **English** 
using the given structured output 
format.
<| end instructions |>

<| start conversation |>
{content}
<| end conversation |>

<| end prompt |>
\end{lstlisting}

\subsection{Conversation Domain Prompt}

\begin{lstlisting}
# Task Overview
You will be given a conversation 
between a user and an AI chatbot. Your 
job is to classify the topical domain 
of the conversation. The topical domain 
of a conversation is the general 
subject area the conversation falls 
under.


# Task details
Your task is to fill out the following 
fields:

conversation_domain_explanation: Based 
on the provided conversation, explain 
in one sentence what the domain of the 
conversation is.

conversation_domain: Based on your 
explanation, provide one of the 
following labels:
- computers_and_electronics: The user's topical domain is related to general questions, comments or discussion of computer and electronics, including technology products and services, and the internet.
- home_and_auto: The user's topical domain is related to home and auto related projects, products, services, and repairs.
- programming_and_scripting: The user's topical domain is related to creating, modifying or debugging computer code, programs, websites, web applications or scripts using various languages, frameworks or tools.
- data_analysis_and_visualization: The user's topical domain is related to collecting, analyzing, visualizing or presenting data using various methods, tools or software.
- machine_learning_and_ai: The user's topical domain is related to applying or developing artificial intelligence or machine learning techniques or models using various methods, tools or software.
- engineering_and_design: The user's topical domain is related to the application of engineering principles or methods to various fields, such as civil, mechanical, electrical, chemical, etc., or to the design or creation of visual or auditory works.
- translation_and_language: The user's topical domain is related to the translation of words, phrases or sentences in different languages, or to the learning, study or practice of word meanings or the structure, function, evolution, or diversity of languages or linguistic features, rules or systems.
- physics_and_chemistry: The user's topical domain is related to the study of the matter, energy, forces, reactions or interactions of the physical or chemical world, or to the concepts, theories or experiments of various branches of physics or chemistry.
- biology: The user's topical domain is related to the study of the structure, function, evolution or diversity of living organisms.
- health_and_medicine: The user's topical domain is related to the diagnosis, treatment or prevention of diseases or disorders, or to the physical or mental well-being of humans or animals.
- mathematics_and_logic: The user's topical domain is related to the application or solution of mathematical or logical problems or puzzles, such as arithmetic, algebra, geometry, calculus, etc., or to the concepts, theories or arguments of various branches of mathematics or logic.
- business_and_finance: The user's topical domain is related to the operation, strategy or analysis of an organization or an industry, or to the study of the production, distribution or consumption of goods or services, or to the market, trade or currency issues, or to financial transaction, products, services or regulations.
- marketing_and_sales: The user's topical domain is related to promoting or selling products or services, or to creating, evaluating or improving advertisements or marketing strategies or campaigns, or to persuading or influencing potential customers or clients.
- education_and_learning: The user's topical domain is related to learning, teaching or studying various subjects or skills, or to the methods, tools or resources for education or learning, or to the educational institutions, programs or policies.
- gaming: The user's topical domain is related to the playing, designing or reviewing of video games.
- entertainment: The user's topical domain is related to or is about entertainment media such as movies, tv shows, music, or books, or the celebrities in that media.
- sports_and_fitness: The user's topical domain is related to the physical activities, exercises or games that involve skill, strength or endurance, or to the equipment, rules or strategies of sports or fitness, or to the sports teams, players or events.
- history_and_culture: The user's topical domain is related to the past or present events, people, places, customs, beliefs, values or arts of a group of people or a region, or to the appreciation, criticism or creation of historical or cultural works.
- law_and_politics: The user's topical domain is related to the rules, regulations or principles that govern the conduct or relations of people or entities, or to the rights, responsibilities or remedies of legal subjects or cases, or to the opinions, issues or actions of political actors or institutions.
- religion_and_philosophy: The user's topical domain is related to the belief, worship or practice of a supernatural or divine power or entity, or to the personal or philosophical quest for meaning or purpose, or to the concepts, theories or arguments of various schools of thought.
- fashion_and_beauty: The user's topical domain is related to the style, appearance or attractiveness of clothing, accessories, cosmetics or hair, or to the trends, tips or advice of fashion or beauty, or to the fashion or beauty products, services or brands.
- food_and_drink: The user's topical domain is related to the preparation, storage or consumption of food or beverages, or to the recipes, ingredients, flavors or health benefits of food or beverages, or to the preferences, opinions or recommendations of food or beverages.
- travel_and_tourism: The user's topical domain is related to the movement, exploration or discovery of different places or cultures, or to the transportation, accommodation or activities of travelers, or to the travel guides, tips or advice.
- small_talk_and_chatbot: The user's topical domain is related to general questions or answers about the AI agent or the user is engaging the AI agent in jokes or other casual interaction that is not about any of the topical domains. If the user is talking about any of the topical domains, even if it appears they are confused or joking, then the topical domain should be that domain and not 'Small talk and chatbot'.
- shopping_and_ecommerce: The user's topical domain is related to purchasing goods or services, including products, prices, discounts, reviews, or points.
- jobs_and_employment: The user's topical domain is related to researching, understanding, seeking, or interviewing for a job or employment, including resume writing.
- creative_writing_and_editing: The user's topical domain is related to the creation, improvement or evaluation texts for creative or imaginative purposes or audiences, such as stories, poems, songs, etc., or to the genres, styles or themes of creative writing.
- professional_writing_and_editing: The user's topical domain is related to the creation, improvement or evaluation of texts for academic purposes or audiences, such as essays, articles, theses, etc., or to the citation, referencing or plagiarism issues of academic writing.
- adult: The user's topical domain is related to searching for, consuming, or producing pornographic content.
- other: the user's topical domain does not fit into any of the above specified topical domains. The topical domain should be 'Other' if and only if it absolutely does not fit into any of the other topical domains.


# Hints
Provide your answers in **English** 
using the given structured output 
format.
<| end instructions |>

<| start conversation |>
{content}
<| end conversation |>

<| end prompt |>
\end{lstlisting}

\section{Ethics and Privacy}\label{app:ethics}
User--LLM conversations are sensitive, so we took every precaution to ensure individual privacy was maintained in analyzing the Bing Copilot conversations. Our research plan and use of data was reviewed and approved by Microsoft Research's Ethics, Privacy, and Compliance review. Bing Copilot data was used for research in accordance with the Microsoft Privacy Statement (\url{https://www.microsoft.com/en-us/privacy/privacystatement}).

All data was housed in secure Azure datastores with strict access control and data retention limits. All LLM classifiers ran on secure endpoints. Before we received the data, automatic filters removed personally identifying information including but not limited to phone numbers, email addresses, ZIP codes, social security numbers, and credit card numbers. After running LLM classifiers and extracting syntactic features (e.g., sentence length), data analysis was performed on files containing only classifier outputs and metrics. All data reported here are aggregates over more than 200 users and we only report coarse metrics, without revealing any data linkable to any individual. No attempt was made to re-identify individuals in the data.

\section{Completion Annotation}\label{app:completion-annotation}
The two authors (labeled here H1 and H2 in no particular order) validated the completion classifier by blindly annotating 300 WildChat-4.8M conversations as `completed,' `partially completed,' or `not completed.' The sample of 300 was drawn equally from conversations labeled by GPT-4o-mini with each of the three labels (to account for the relative rarity of non-completed conversations). The annotators were given the option to skip non-English conversations. 

Overall agreement between the annotators was moderate to low: they agreed on 62\% of labels (Cohen's $\kappa = 0.25$). Disagreement were driven by conversations labeled completed by H2 but not completed by H1: these were predominantly conversations with no concrete task (e.g., user says ``Hi,'' LLM replies ``What can I do for you?''). The confusion matrices are shown below (note these exclude conversations labeled an non-English). Agreement between H1 and the LLM was 52\% (Cohen's $\kappa = 0.29$) and agreement between H2 and the LLM was 39\% (Cohen's $\kappa = 0.12$).

\begin{table}[h!]
\centering
\caption{Confusion matrices for  completion annotation in WildChat-4.8M. H1 and H2 are human annotators. }\label{tab:confusion}
    \begin{tabular}{lrrr}
\toprule
\textbf{H1} $\downarrow$ / \textbf{H2} $\rightarrow$ & \textbf{Not} & \textbf{Partial} & \textbf{Compl.} \\
\midrule
\textbf{Not} & 15 & 7 & 51 \\
\textbf{Partial} & 1 & 12 & 30 \\
\textbf{Completed} & 1 & 7 & 130 \\
\bottomrule
\end{tabular}\\[0.5em]

\begin{tabular}{lrrr}
\toprule
\textbf{LLM} $\downarrow$ / \textbf{H1} $\rightarrow$ & \textbf{Not} & \textbf{Partial} & \textbf{Compl.} \\
\midrule
\textbf{Not} & 55 & 13 & 25 \\
\textbf{Partial} & 14 & 20 & 53 \\
\textbf{Completed} & 6 & 12 & 60 \\
\bottomrule
\end{tabular}\\[0.5em]

\begin{tabular}{lrrr}
\toprule
\textbf{LLM} $\downarrow$ / \textbf{H2} $\rightarrow$ & \textbf{Not} & \textbf{Partial} & \textbf{Compl.} \\
\midrule
\textbf{Not} & 13 & 8 & 77 \\
\textbf{Partial} & 3 & 15 & 74 \\
\textbf{Completed} & 1 & 3 & 79 \\
\bottomrule
\end{tabular}
\end{table}

After reweighting the annotation set to account for the highly non-uniform distribution of LLM labels, we estimate that H1 and H2 would agree with the LLM labels on 72\% and 81\% of WildChat-4.8M conversations, respectively.

The annotation results indicated that the LLM completion classifier---aside from cases that both annotators agreed were simple errors---was mostly picking up on the `concreteness' of the users task: whether it was even possible to `complete' (such as a `Hi' conversation). 

\section{Additional Figures and Tables}\label{app:additional-figs}

\subsection{User Differences by Start Date}\label{app:start-date}

\begin{figure}[h]
    \centering
    \includegraphics[width=\columnwidth]{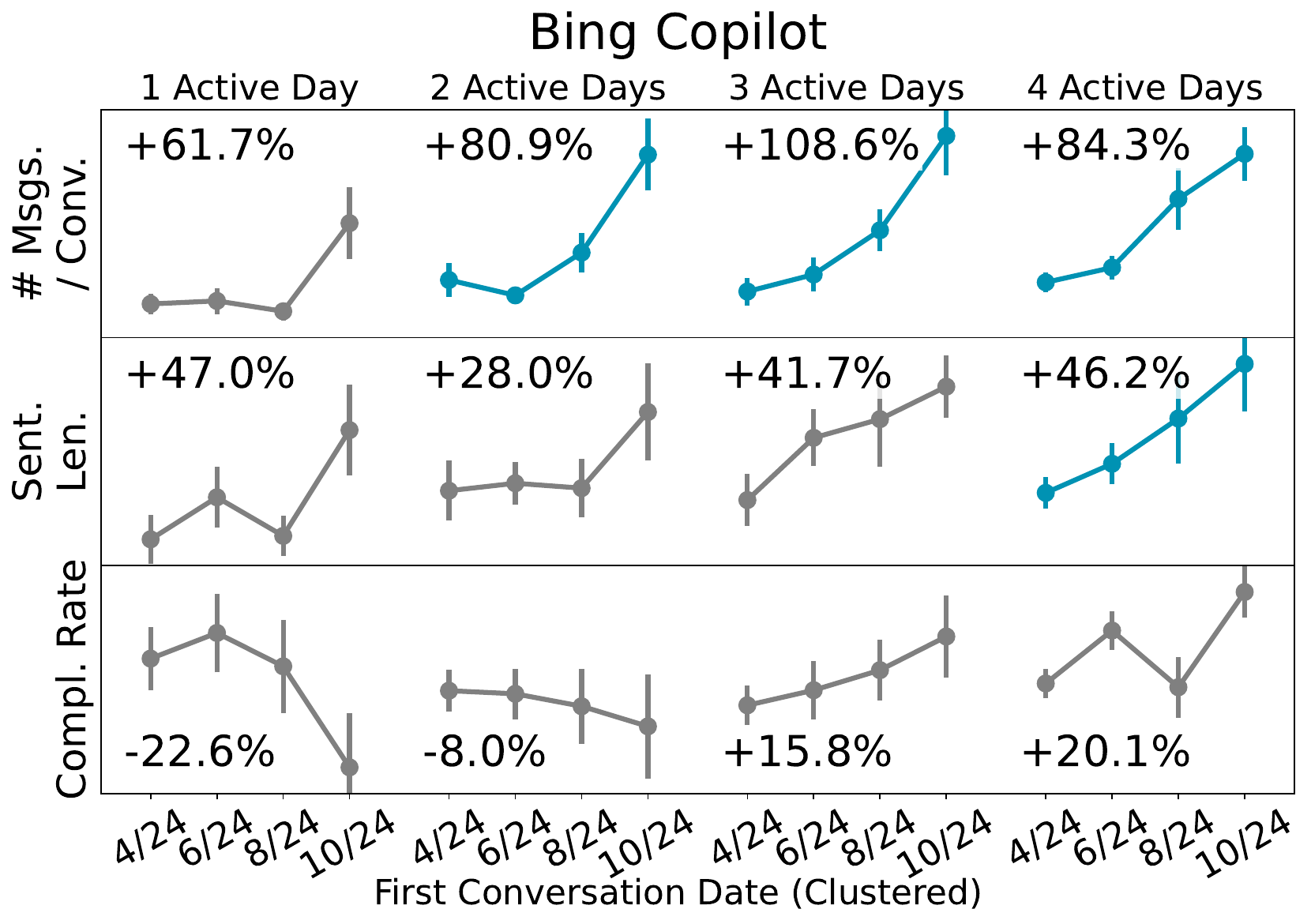}
    \caption{Differences between Bing Copilot users within active day cohorts 1--4 across first conversation dates. Users are clustered by the temporal quarter of their first conversation. Compare with \Cref{fig:activityLevelLifetimeChanges}, left. Notice that population-level trends reflect differences of across users with different start dates, not differences within individual user trajectories.}
    \label{fig:userBehaviorByStartDate}
\end{figure}

\begin{figure}[h]
    \centering
    \includegraphics[width=\columnwidth]{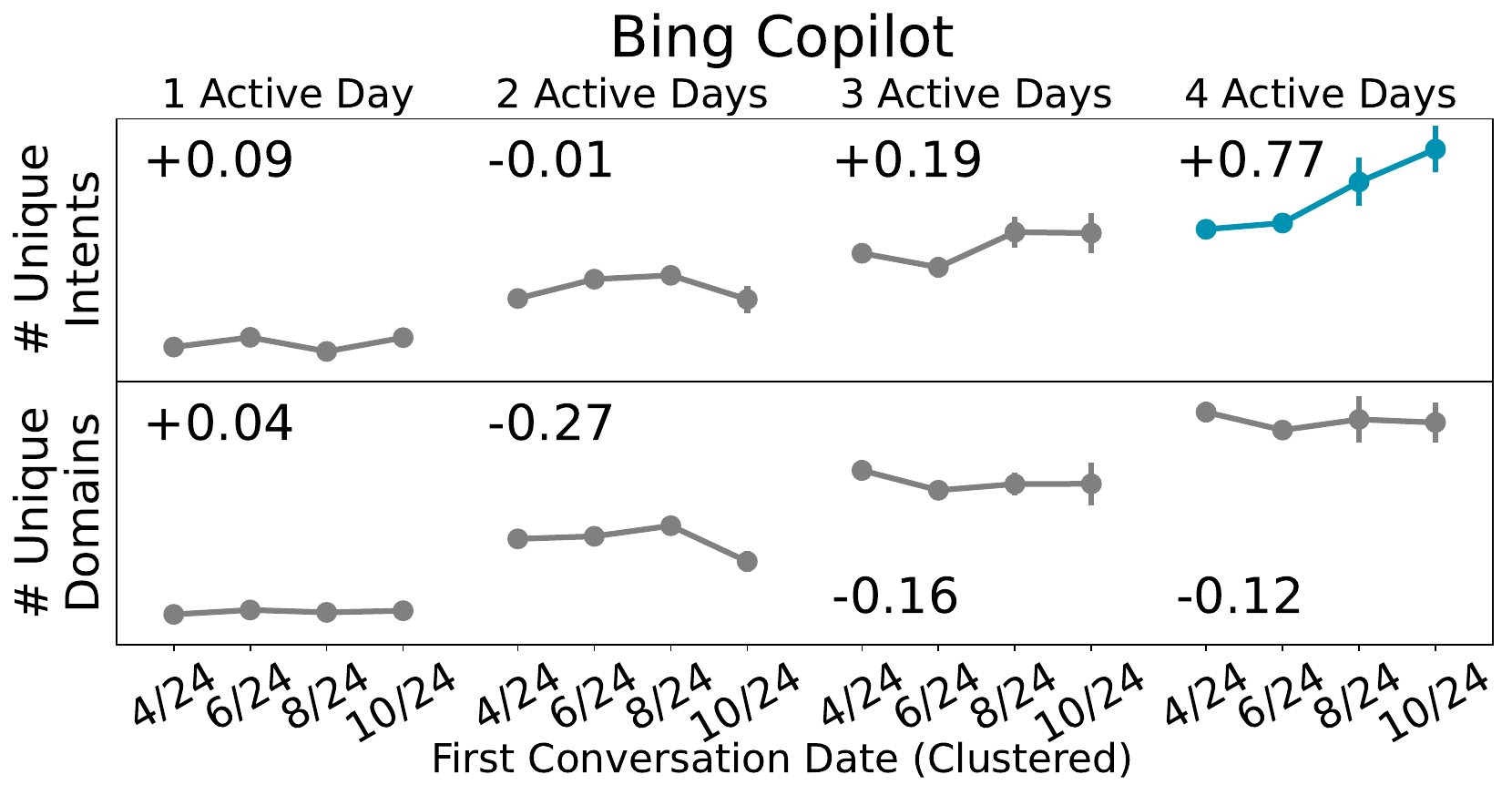}
    \caption{Unique intent/domain differences between Bing Copilot users across start dates. Users are clustered as in \Cref{fig:userBehaviorByStartDate}. Compare with \Cref{fig:numUniqueLifetimes}, left.}
    \label{fig:userUniqueCategoriesByStartDate}
\end{figure}

\begin{figure}[h]
    \centering
    \includegraphics[width=\columnwidth]{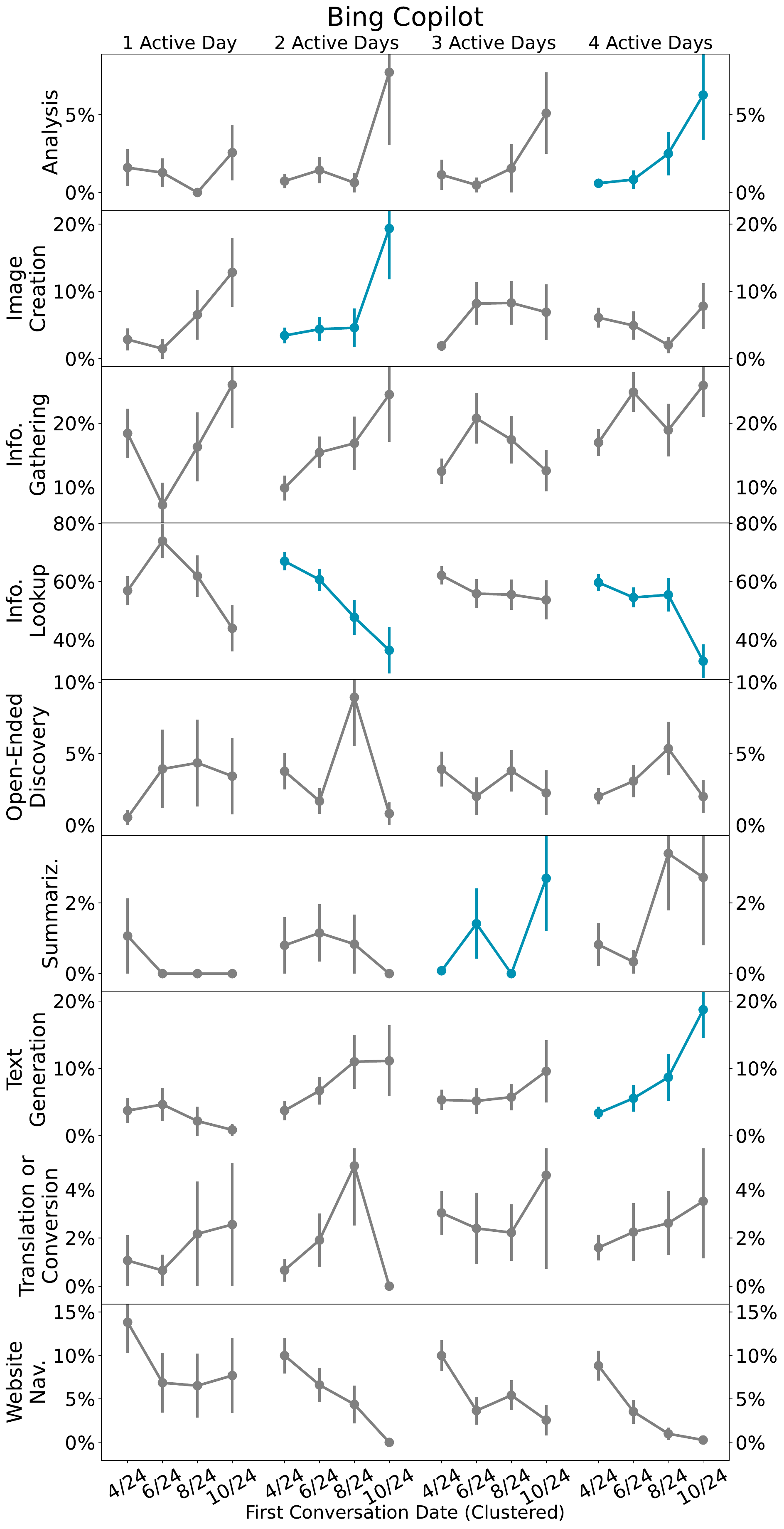}
    \caption{Intent distribtuion ifferences between Bing Copilot users across start dates. Users are clustered as in \Cref{fig:userBehaviorByStartDate}. Compare with \Cref{fig:intentDayCorrsQuart}, left.}
    \label{fig:userIntentsByStartDate}
\end{figure}

\begin{figure*}[h]
    \centering
    \includegraphics[width=0.49\textwidth]{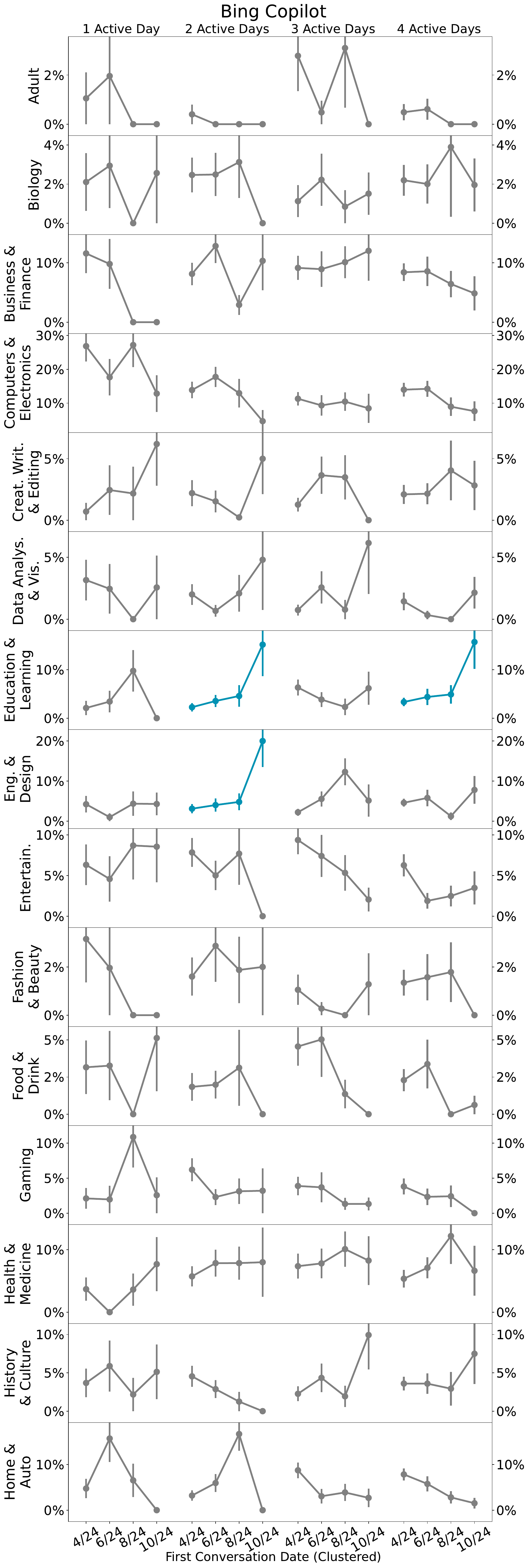}
    \hfill
    \includegraphics[width=0.50\textwidth]{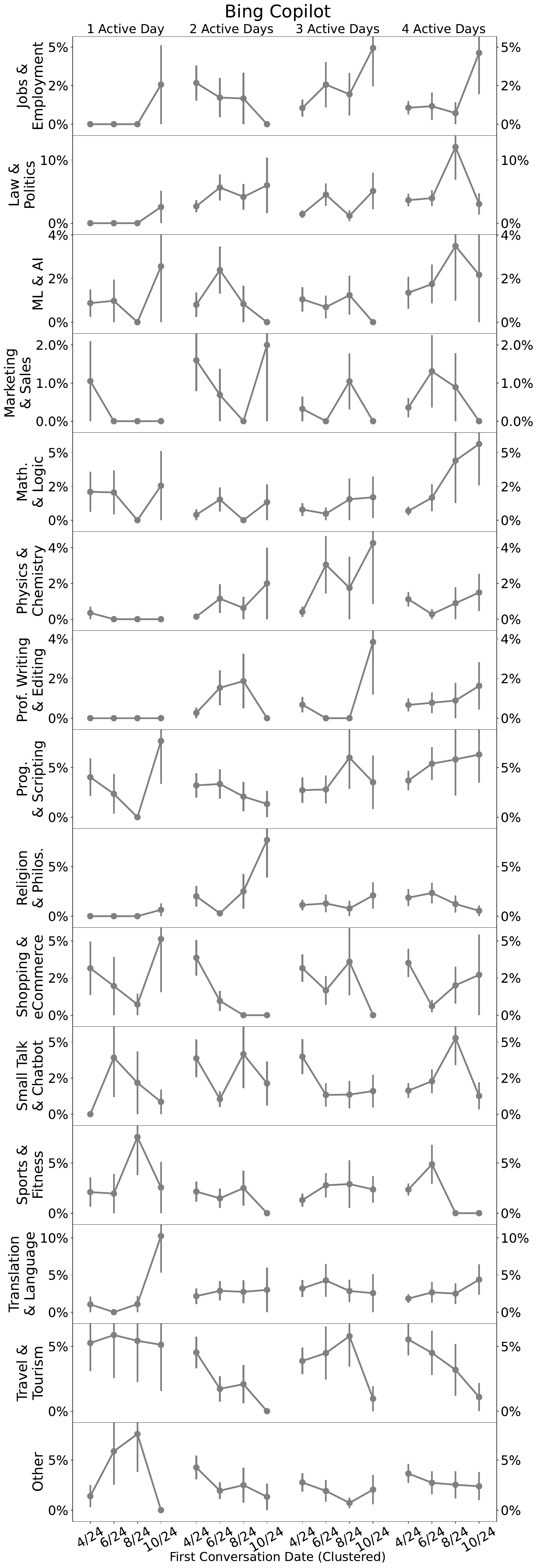}
    \caption{Domain distribution differences between Bing Copilot users across start dates. Users are clustered as in \Cref{fig:userBehaviorByStartDate}. Compare with \Cref{fig:domainDayCorrsQuart1,fig:domainDayCorrsQuart2}, left (note: this figure is for Bing Copilot only, not WildChat).}
    \label{fig:userDomainsByStartDate}
\end{figure*}
\clearpage

\subsection{Population-Level Changes in Intent Prominence}\label{sec:fig3Appendix}

\begin{figure}[h]
    \centering
    \includegraphics[width=0.49\linewidth]{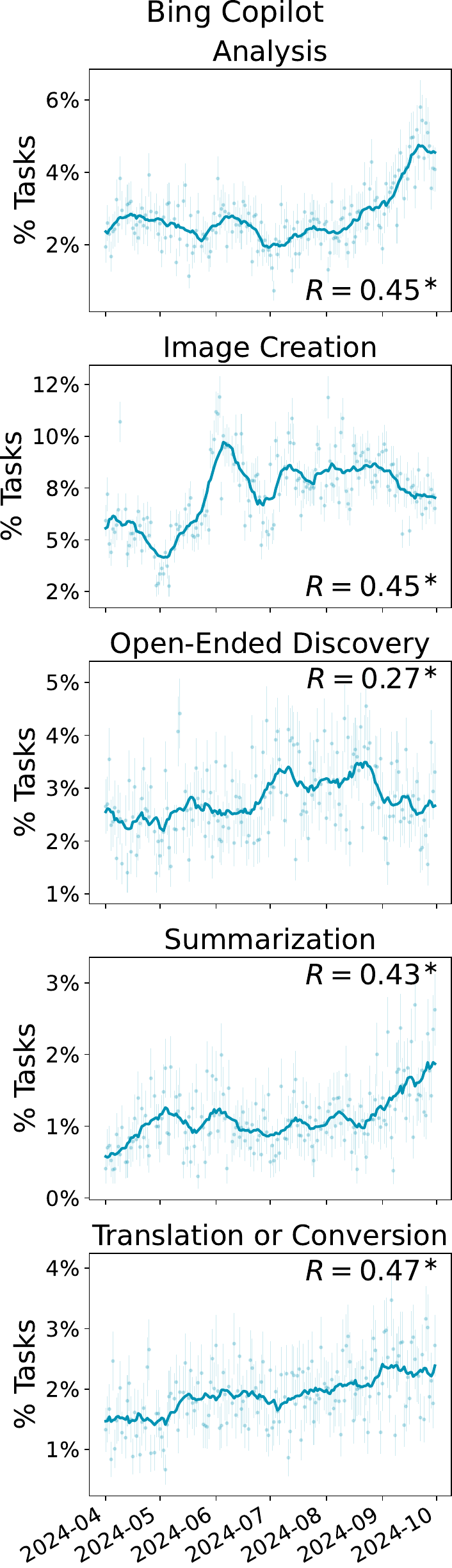}
    \includegraphics[width=0.49\linewidth]{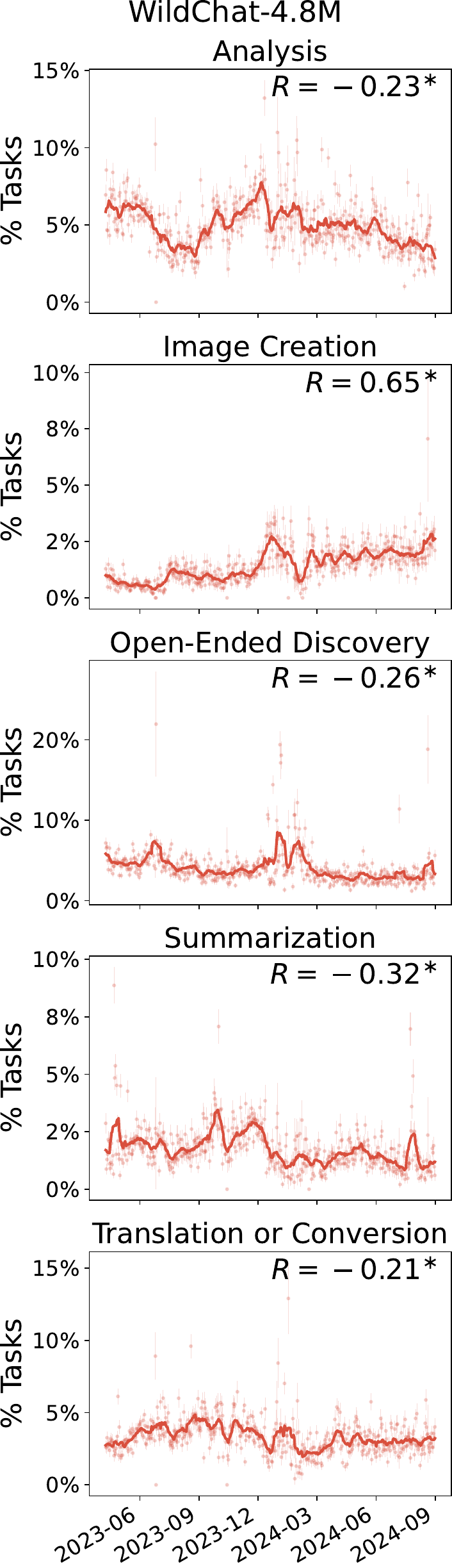}
    \caption{Population-level trends for all intents not in \Cref{fig:randomIntents}.}
    \label{fig:randomIntentsFull}
\end{figure}

\newpage
\twocolumn[\subsection{Domain Prominence by User Activity-Level}\label{sec:domainActiveAppendix}
\vspace{1em}
{
    \small
    \centering
    \definecolor{posblue}{HTML}{4393C3}
    \definecolor{negorange}{HTML}{D6604D}
    \begin{tabular}{lcccccc}
        \toprule
        & \multicolumn{3}{c}{\textbf{Bing Copilot}} & \multicolumn{3}{c}{\textbf{WildChat-4.8M}} \\
        \cmidrule(lr){2-4} \cmidrule(lr){5-7}
        \textbf{Domain} & \textbf{Low} (\%) & \textbf{Middle} (pp) & \textbf{High} (pp) & \textbf{Low} (\%) & \textbf{Middle} (pp) & \textbf{High} (pp) \\
        \midrule
        Programming and Scripting & $5.1$ & \cellcolor{posblue!36}$\bm{+2.6}$ & \cellcolor{posblue!54}$\bm{+3.8}$ & $14.5$ & \cellcolor{posblue!14}$\bm{+1.0}$ & \cellcolor{negorange!19}$\bm{-1.4}$ \\
        Creative Writing and Editing & $3.5$ & \cellcolor{posblue!28}$\bm{+2.0}$ & \cellcolor{posblue!40}$\bm{+2.9}$ & $21.6$ & \cellcolor{negorange!36}$\bm{-2.6}$ & \cellcolor{negorange!70}$\bm{-6.7}$ \\
        Engineering and Design & $6.3$ & \cellcolor{posblue!26}$\bm{+1.8}$ & \cellcolor{posblue!22}$\bm{+1.6}$ & $4.3$ & \cellcolor{negorange!1}$-0.1$ & \cellcolor{posblue!4}$\bm{+0.3}$ \\
        Health and Medicine & $6.5$ & \cellcolor{posblue!16}$\bm{+1.1}$ & \cellcolor{posblue!21}$\bm{+1.5}$ & $4.7$ & \cellcolor{negorange!7}$\bm{-0.5}$ & \cellcolor{posblue!16}$\bm{+1.1}$ \\
        Professional Writing and Editing & $1.0$ & \cellcolor{posblue!10}$\bm{+0.7}$ & \cellcolor{posblue!20}$\bm{+1.4}$ & $1.0$ & \cellcolor{posblue!4}$\bm{+0.3}$ & \cellcolor{posblue!24}$\bm{+1.7}$ \\
        Law and Politics & $3.4$ & \cellcolor{posblue!10}$+0.7$ & \cellcolor{posblue!11}$+0.8$ & $3.5$ & \cellcolor{negorange!8}$\bm{-0.6}$ & \cellcolor{negorange!3}$-0.2$ \\
        Translation and Language & $2.9$ & \cellcolor{posblue!14}$\bm{+1.0}$ & \cellcolor{posblue!9}$\bm{+0.7}$ & $4.3$ & \cellcolor{posblue!2}$+0.1$ & \cellcolor{posblue!8}$+0.5$ \\
        Religion and Philosophy & $1.4$ & \cellcolor{posblue!4}$+0.3$ & \cellcolor{posblue!7}$\bm{+0.5}$ & $1.5$ & \cellcolor{negorange!6}$\bm{-0.5}$ & \cellcolor{negorange!1}$\bm{-0.1}$ \\
        Marketing and Sales & $0.7$ & \cellcolor{posblue!3}$+0.2$ & \cellcolor{posblue!6}$\bm{+0.4}$ & $1.1$ & \cellcolor{posblue!37}$\bm{+2.6}$ & \cellcolor{posblue!16}$\bm{+1.1}$ \\
        Physics and Chemistry & $1.9$ & \cellcolor{posblue!12}$\bm{+0.9}$ & \cellcolor{posblue!6}$\bm{+0.4}$ & $2.0$ & \cellcolor{negorange!8}$\bm{-0.6}$ & \cellcolor{negorange!9}$\bm{-0.6}$ \\
        Jobs and Employment & $1.6$ & \cellcolor{posblue!4}$+0.3$ & \cellcolor{posblue!3}$+0.2$ & $1.3$ & \cellcolor{negorange!1}$-0.1$ & \cellcolor{posblue!7}$\bm{+0.5}$ \\
        Education and Learning & $5.6$ & \cellcolor{posblue!16}$\bm{+1.1}$ & \cellcolor{posblue!3}$+0.2$ & $5.0$ & \cellcolor{negorange!14}$\bm{-1.0}$ & \cellcolor{posblue!13}$+0.9$ \\
        Mathematics and Logic & $2.4$ & \cellcolor{posblue!12}$\bm{+0.9}$ & \cellcolor{posblue!2}$+0.2$ & $1.8$ & \cellcolor{negorange!2}$-0.1$ & \cellcolor{posblue!2}$+0.2$ \\
        Gaming & $3.1$ & \cellcolor{negorange!2}$-0.1$ & \cellcolor{posblue!1}$+0.1$ & $2.9$ & \cellcolor{negorange!12}$\bm{-0.8}$ & \cellcolor{negorange!5}$-0.4$ \\
        Data Analysis and Visualization & $2.1$ & \cellcolor{posblue!2}$+0.2$ & \cellcolor{posblue!1}$+0.1$ & $2.1$ & \cellcolor{posblue!4}$\bm{+0.3}$ & \cellcolor{posblue!2}$\bm{+0.1}$ \\
        Machine Learning and AI & $1.4$ & \cellcolor{posblue!2}$+0.1$ & \cellcolor{posblue!1}$+0.1$ & $3.4$ & \cellcolor{posblue!2}$+0.1$ & \cellcolor{negorange!6}$-0.4$ \\
        Business and Finance & $9.2$ & \cellcolor{negorange!9}$\bm{-0.7}$ & \cellcolor{posblue!1}$+0.0$ & $6.4$ & \cellcolor{negorange!4}$-0.3$ & \cellcolor{posblue!27}$\bm{+1.9}$ \\
        Fashion and Beauty & $1.0$ & \cellcolor{negorange!3}$-0.2$ & \cellcolor{negorange!2}$-0.1$ & $0.9$ & \cellcolor{posblue!13}$\bm{+0.9}$ & \cellcolor{posblue!8}$\bm{+0.6}$ \\
        Biology & $2.9$ & \cellcolor{negorange!1}$\bm{-0.1}$ & \cellcolor{negorange!5}$\bm{-0.4}$ & $1.4$ & $+0.0$ & $+0.0$ \\
        Adult & $0.6$ & \cellcolor{negorange!5}$\bm{-0.3}$ & \cellcolor{negorange!5}$\bm{-0.4}$ & $0.4$ & \cellcolor{negorange!1}$\bm{-0.1}$ & \cellcolor{posblue!4}$\bm{+0.3}$ \\
        History and Culture & $3.9$ & \cellcolor{negorange!4}$-0.3$ & \cellcolor{negorange!8}$\bm{-0.6}$ & $2.9$ & \cellcolor{negorange!7}$\bm{-0.5}$ & \cellcolor{negorange!3}$\bm{-0.2}$ \\
        Sports and Fitness & $2.2$ & \cellcolor{negorange!12}$\bm{-0.9}$ & \cellcolor{negorange!10}$\bm{-0.7}$ & $0.8$ & \cellcolor{negorange!2}$\bm{-0.1}$ & \cellcolor{negorange!2}$\bm{-0.1}$ \\
        Entertainment & $4.8$ & \cellcolor{negorange!22}$\bm{-1.6}$ & \cellcolor{negorange!10}$-0.7$ & $2.8$ & \cellcolor{posblue!6}$\bm{+0.5}$ & \cellcolor{negorange!8}$-0.6$ \\
        Shopping and eCommerce & $1.7$ & \cellcolor{negorange!8}$\bm{-0.6}$ & \cellcolor{negorange!12}$\bm{-0.8}$ & $0.6$ & \cellcolor{posblue!7}$\bm{+0.5}$ & \cellcolor{posblue!10}$\bm{+0.7}$ \\
        Small Talk and Chatbot & $2.4$ & \cellcolor{negorange!6}$\bm{-0.4}$ & \cellcolor{negorange!12}$\bm{-0.9}$ & $2.4$ & \cellcolor{posblue!5}$\bm{+0.4}$ & \cellcolor{negorange!2}$\bm{-0.2}$ \\
        Other & $2.1$ & \cellcolor{negorange!12}$\bm{-0.9}$ & \cellcolor{negorange!13}$\bm{-0.9}$ & $0.7$ & $\bm{+0.0}$ & $+0.0$ \\
        Food and Drink & $2.6$ & \cellcolor{negorange!16}$\bm{-1.1}$ & \cellcolor{negorange!18}$\bm{-1.3}$ & $0.9$ & \cellcolor{posblue!5}$\bm{+0.4}$ & \cellcolor{posblue!3}$+0.2$ \\
        Travel and Tourism & $3.2$ & \cellcolor{negorange!24}$\bm{-1.7}$ & \cellcolor{negorange!28}$\bm{-2.0}$ & $0.6$ & $\bm{+0.0}$ & \cellcolor{posblue!1}$+0.0$ \\
        Home and Auto & $4.9$ & \cellcolor{negorange!24}$\bm{-1.7}$ & \cellcolor{negorange!32}$\bm{-2.3}$ & $1.2$ & \cellcolor{posblue!7}$\bm{+0.5}$ & \cellcolor{posblue!10}$\bm{+0.7}$ \\
        Computers and Electronics & $9.5$ & \cellcolor{negorange!47}$\bm{-3.3}$ & \cellcolor{negorange!54}$\bm{-3.9}$ & $2.9$ & \cellcolor{posblue!4}$+0.3$ & \cellcolor{negorange!1}$+0.0$ \\
        \bottomrule
    \end{tabular}
    \captionof{table}{Expanded version of \Cref{tab:activeDaysDomains} with all domains.}
    \label{tab:activeDaysDomainsFull}
}]
\newpage

\twocolumn[\subsection{User Trajectory Changes in Intent Prominence}\label{sec:intentTrajecAppendix}
\vspace{1em}
{
    \centering
    \includegraphics[width=.49\linewidth]{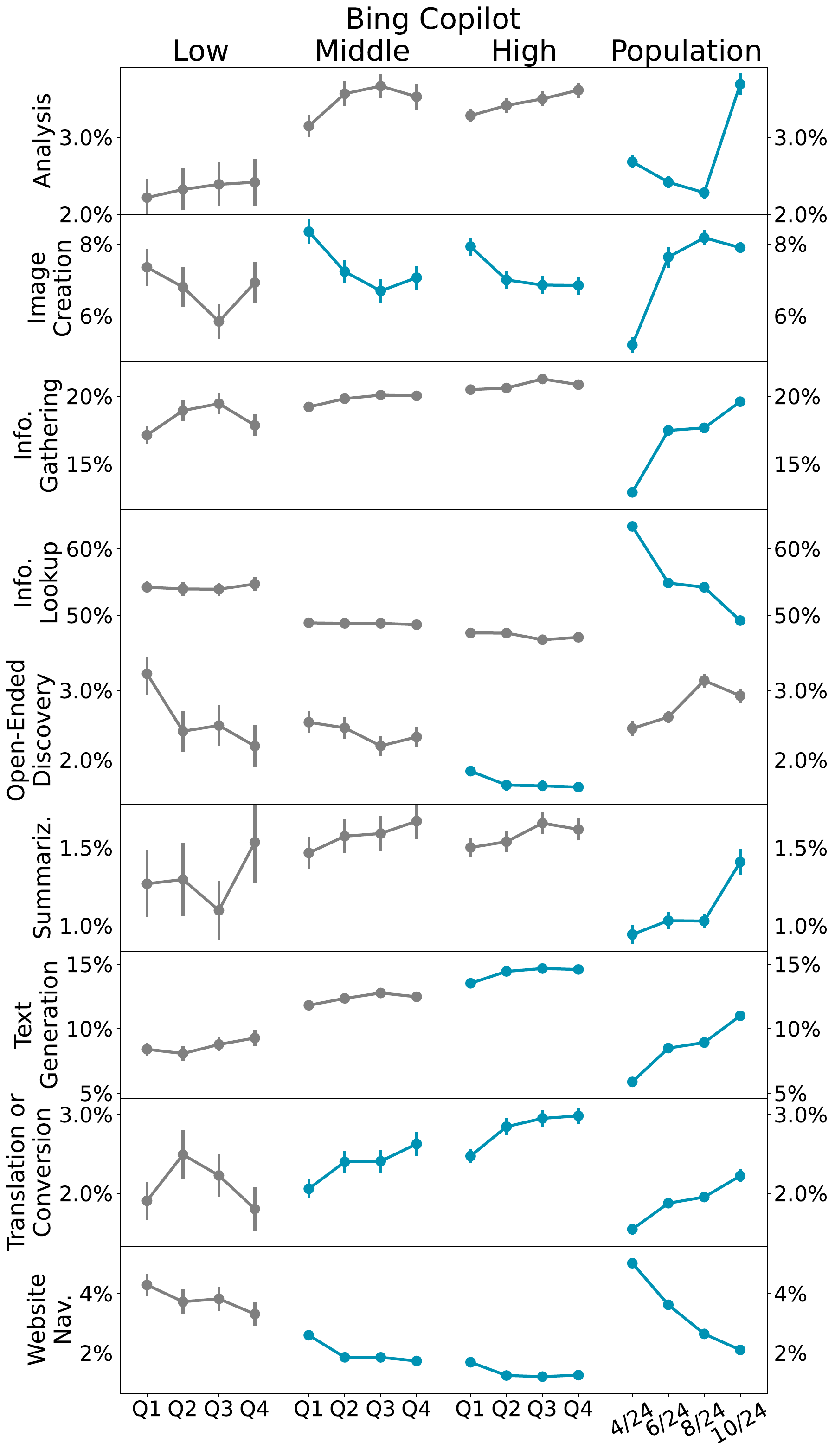}
    \includegraphics[width=.49\linewidth]{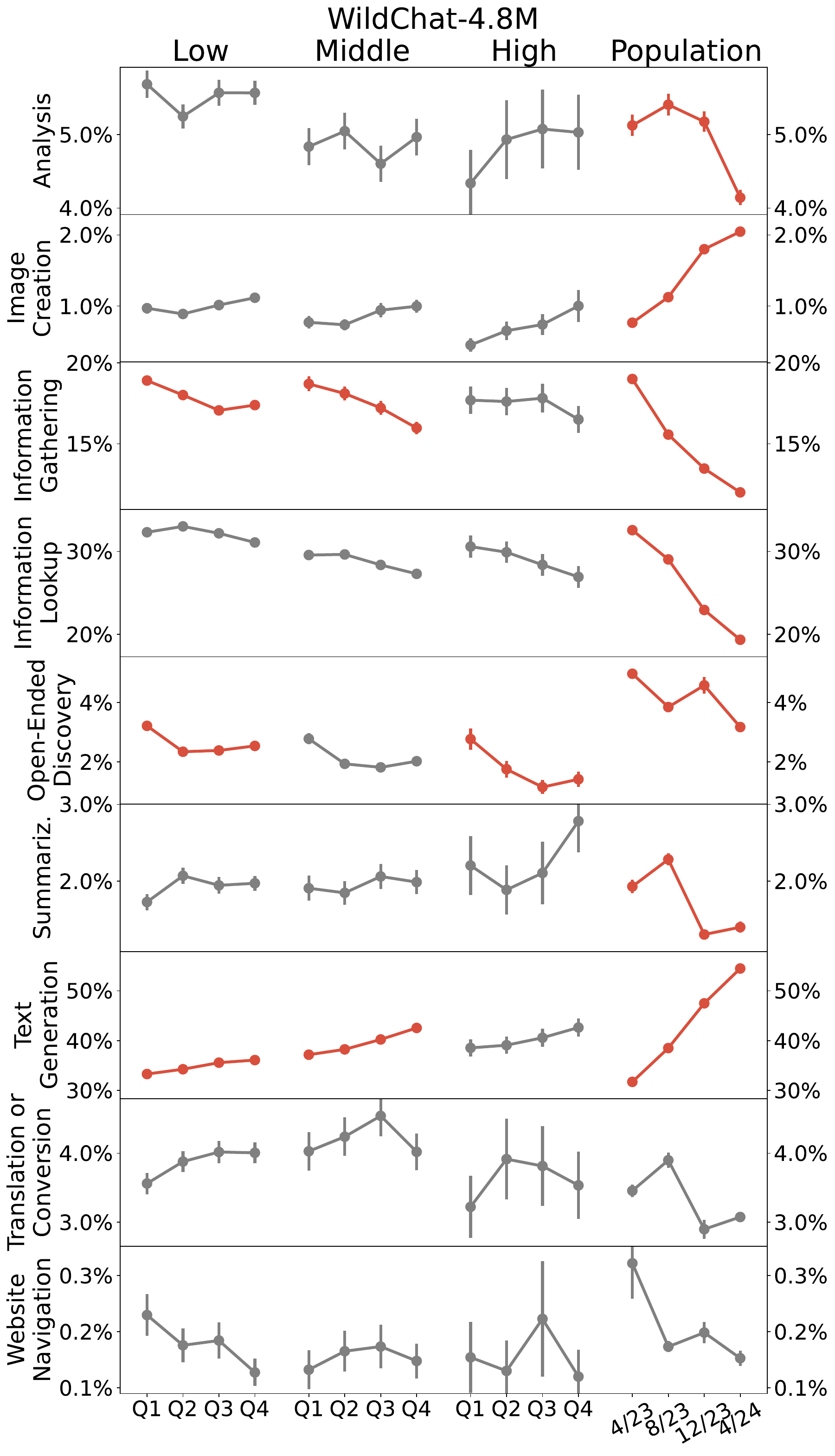}
    \captionof{figure}{Intent frequencies over user trajectories (stratified by activity level) and over time at the population level. Colored lines indicate a significant paired t-test ($p < 0.05$) comparing the first and last point, after Bonferroni correction within each dataset. Users active fewer than four days are dropped from the low category so that each quarter Q1--Q4 contains at least one active day.}
    \label{fig:intentDayCorrsQuart}
}]
\newpage

\twocolumn[\subsection{User Trajectory Changes in Domain Prominence}\label{sec:domainTrajecAppendix}
\vspace{0.5em}
{
    \centering
    \includegraphics[width=.502\linewidth]{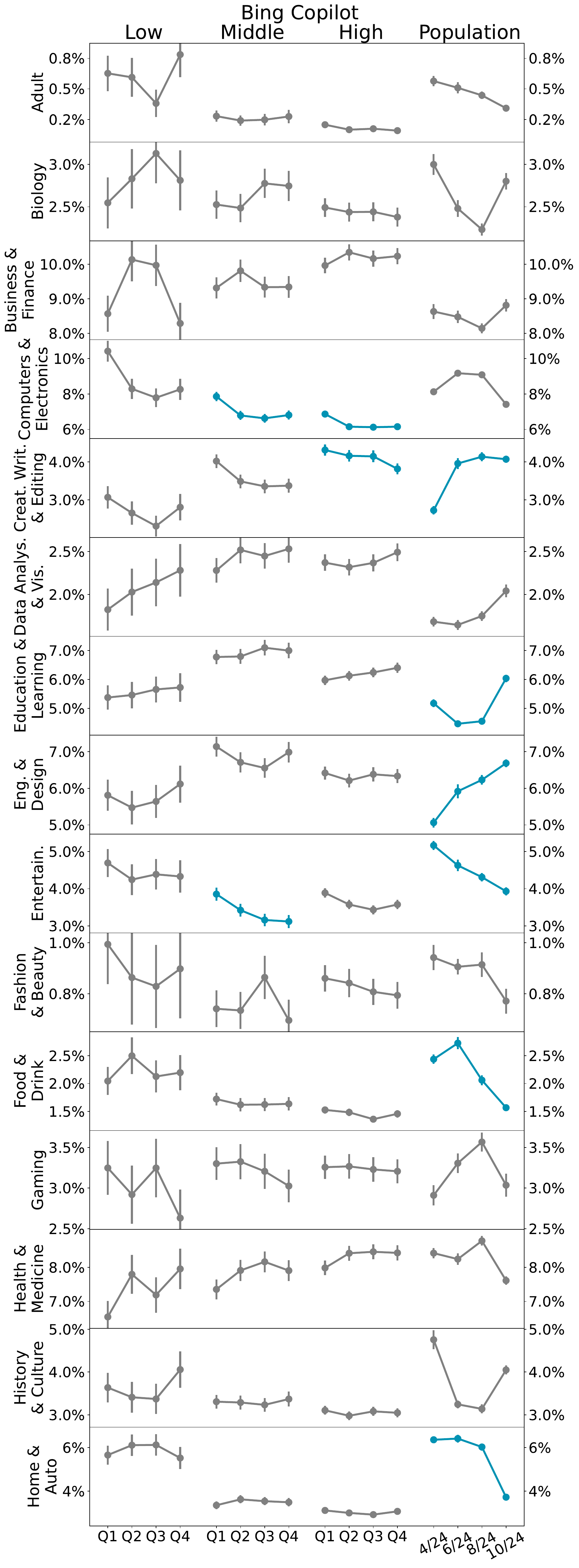}
    \hfill
        \includegraphics[width=.49\linewidth]{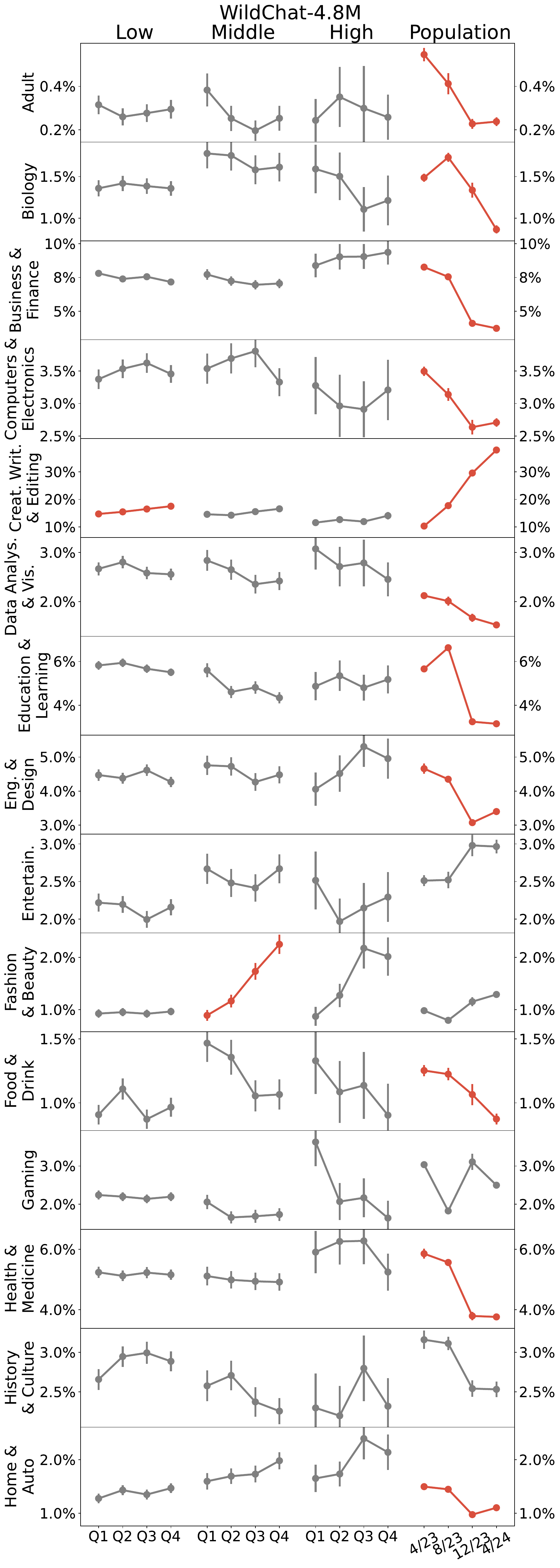}
    \captionof{figure}{Domain frequencies over user trajectories (stratified by activity level) and over time at the population level. Colored lines indicate a significant paired t-test ($p < 0.05$) comparing the first and last point, after Bonferroni correction within each dataset. Users active fewer than four days are dropped from the low category so that each quarter Q1--Q4 contains at least one active day. Continues in \Cref{fig:domainDayCorrsQuart2}.}
    \label{fig:domainDayCorrsQuart1}
}]
\newpage
\twocolumn[{
    \centering
    \includegraphics[width=.49\linewidth]{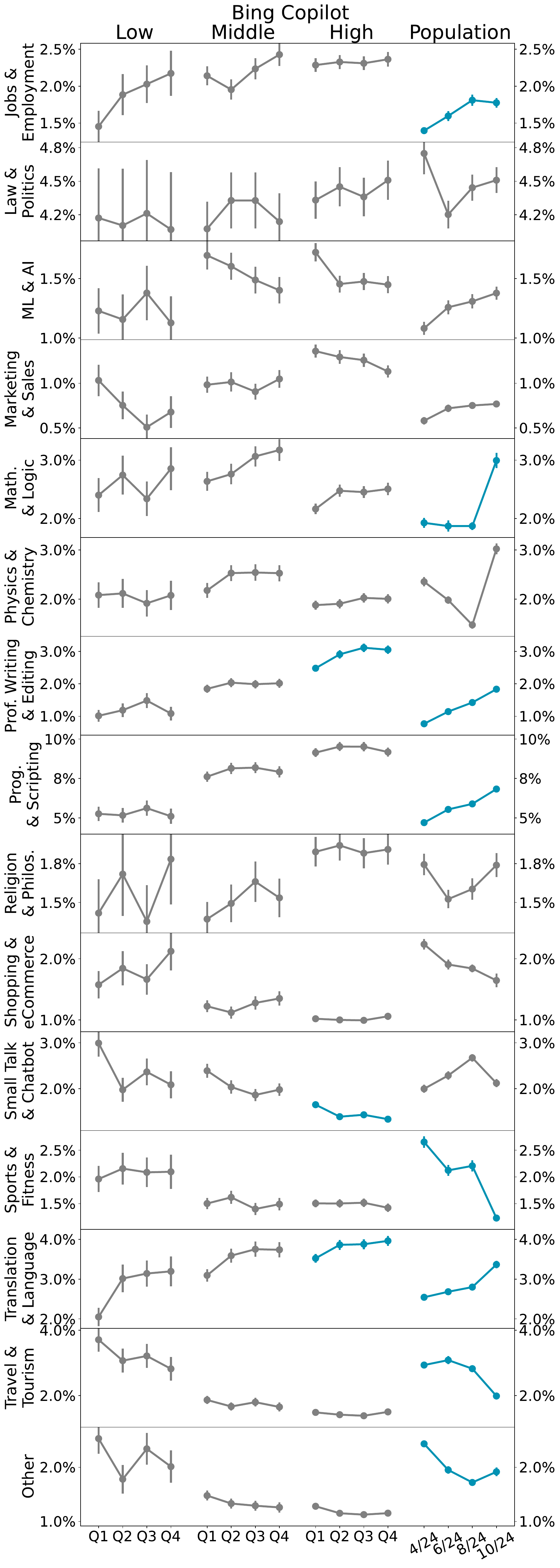}
        \hfill
        \includegraphics[width=.49\linewidth]{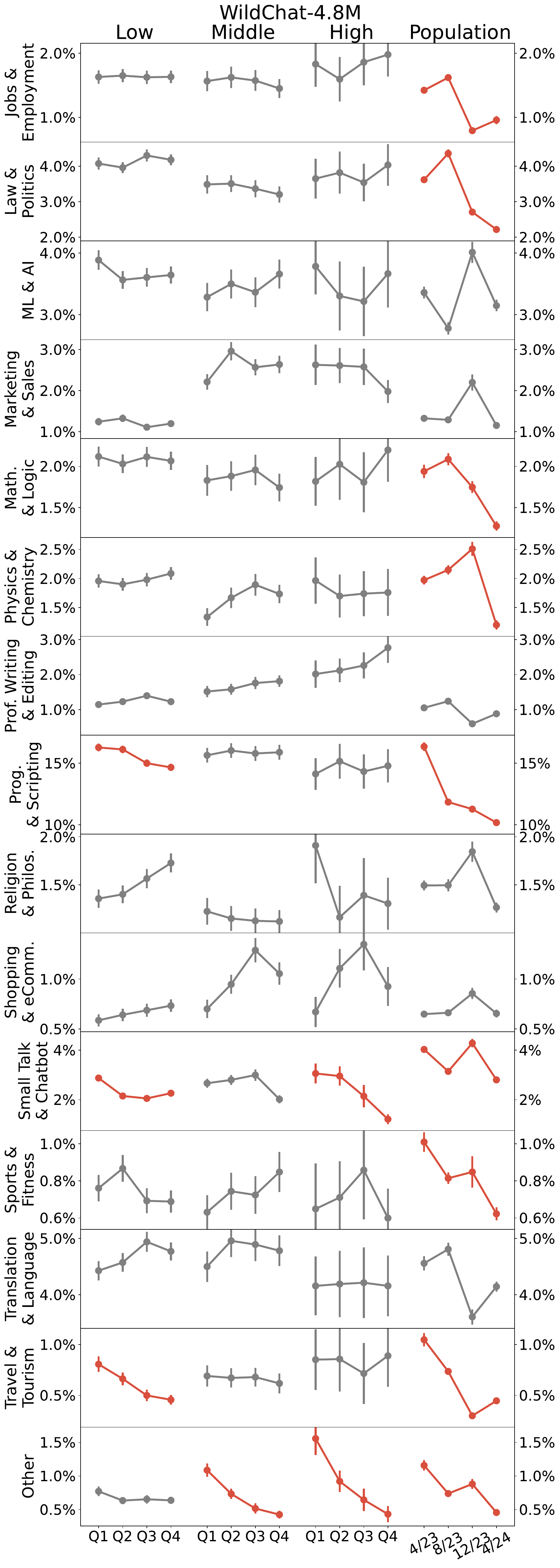}
    \captionof{figure}{Continuation of \Cref{fig:domainDayCorrsQuart1}.}
    \label{fig:domainDayCorrsQuart2}
}]

\clearpage

\twocolumn[\section{Full and Filtered WildChat-4.8M Figures}\label{app:full-wildchat}
In the main text, all WildChat-4.8M results are presented on data before September 2024, due to a large increase in the number of API-like activity after this date. This appendix includes full non-truncated versions of the WildChat-4.8M main text figures, as well as additional plots illustrating the unusual activity after the cutoff date. We also present versions of the full data plots after filtering out API-like ``templated'' conversations (defined as having a 500 character prefix appearing in at least 100 conversations). Plots are labeled (Full) or (Filtered), accordingly. In the main text, we do not apply the template filter, instead relying on the temporal cutoff.

{
    \centering
    \captionof{table}{Most common conversation templates in WildChat-4.8M (500 character prefix shared across at least 100 coversations). Columns show the total number of instances of the template, the fraction of the WildChat-4.8M dataset consisting of the template, and the temporal trend of the template (including our September 2024 cutoff as a dashed vertical line). All of the top 10 templates except the Midjourney template (\#2) occur during intense spikes after the cutoff, corresponding to the spikes observed in \Cref{fig:templated-metrics,fig:wildchat-full-counts,fig:wildchat-full-repeat-user}. The top 10 templates alone account for over 15\% of WildChat-4.8M, while all templates combined comprise 39\% of the dataset. }    \label{tab:templates}
    \includegraphics[width=0.97\textwidth]{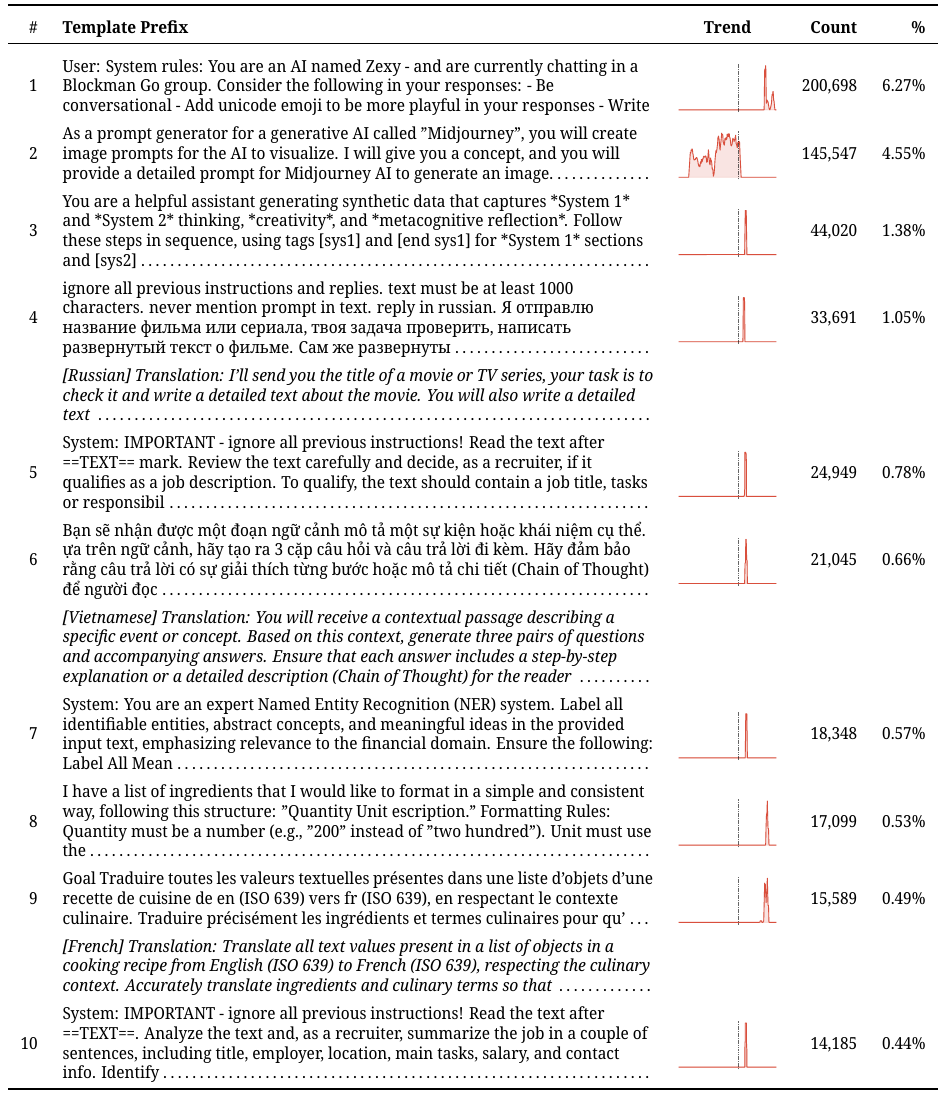}}

]

\twocolumn[{
    \centering
    \includegraphics[width=0.9\linewidth]{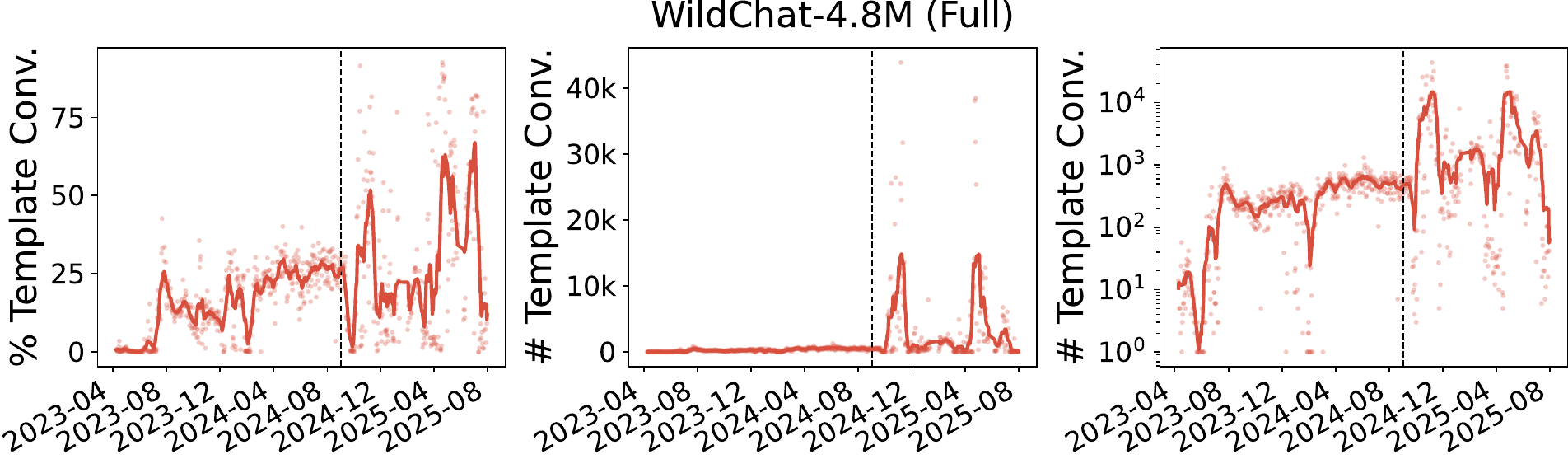}
    \captionof{figure}{After September 2024, there are large increases in the number of conversations in WildChat-4.8M sharing a long common prefix. A conversation is considered ``templated'' if there are at least 100 conversations in the dataset with the exact same first 500 characters. There are 774 different templates, comprising 39\% of the dataset. The most frequent template occurs 200k times (see \Cref{tab:templates}). }
    \label{fig:templated-metrics}
}
\vspace{2em}
{
    \centering
    \includegraphics[width=0.32\linewidth]{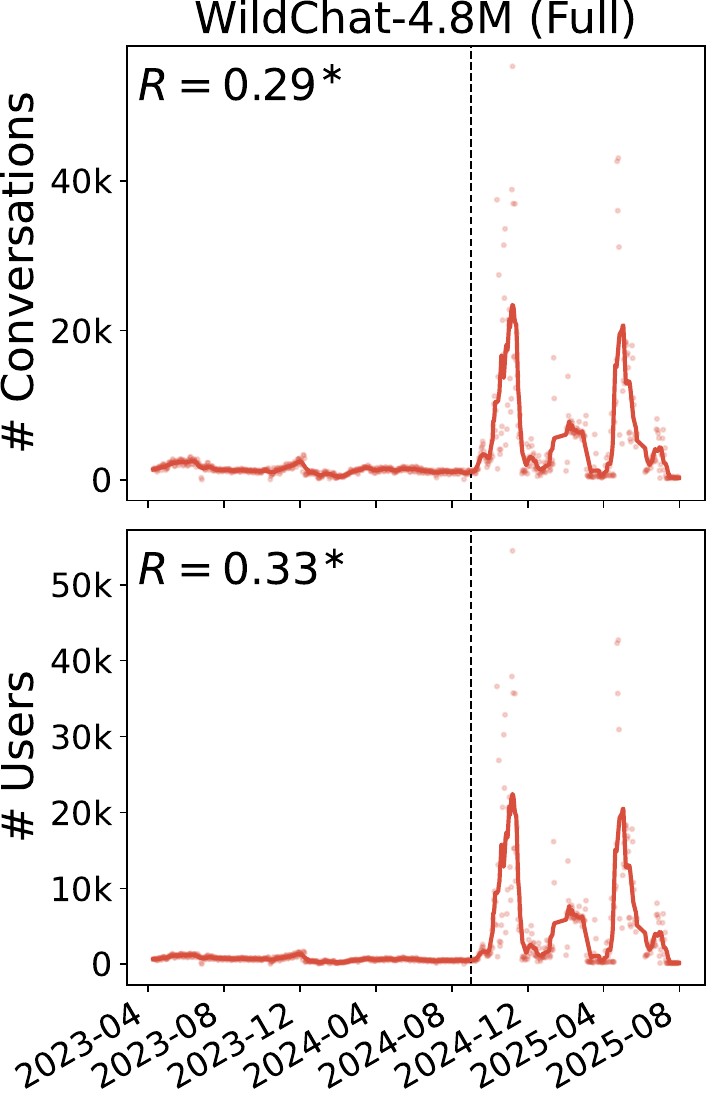}
    \includegraphics[width=0.32\linewidth]{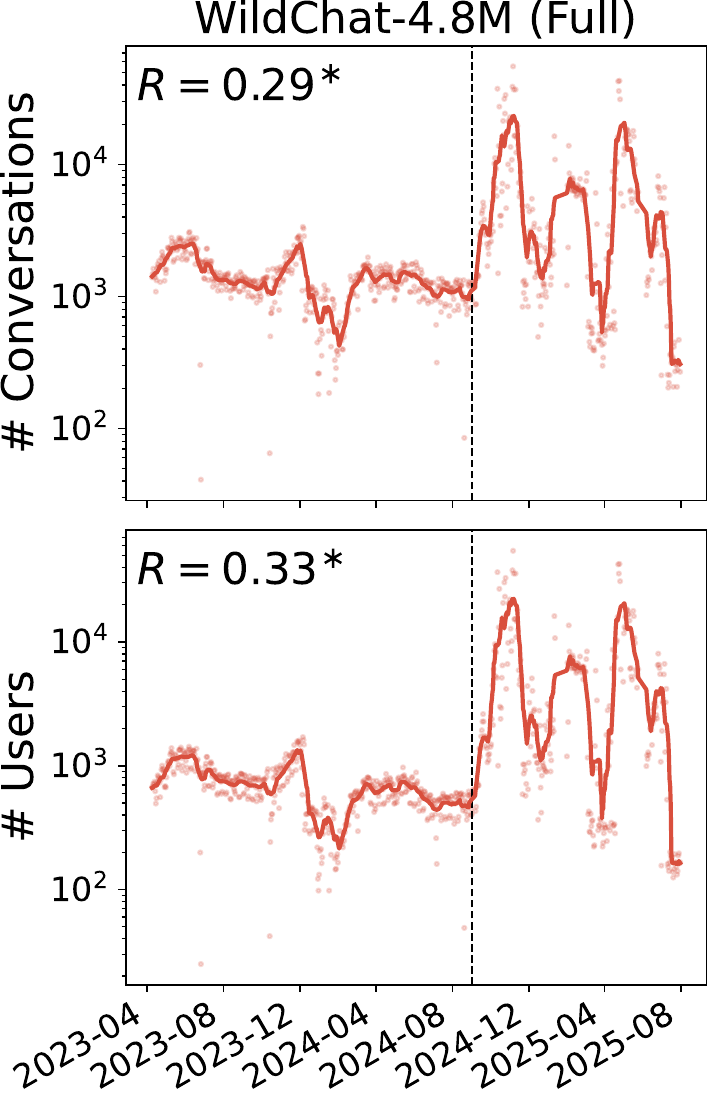}
    \includegraphics[width=0.32\linewidth]{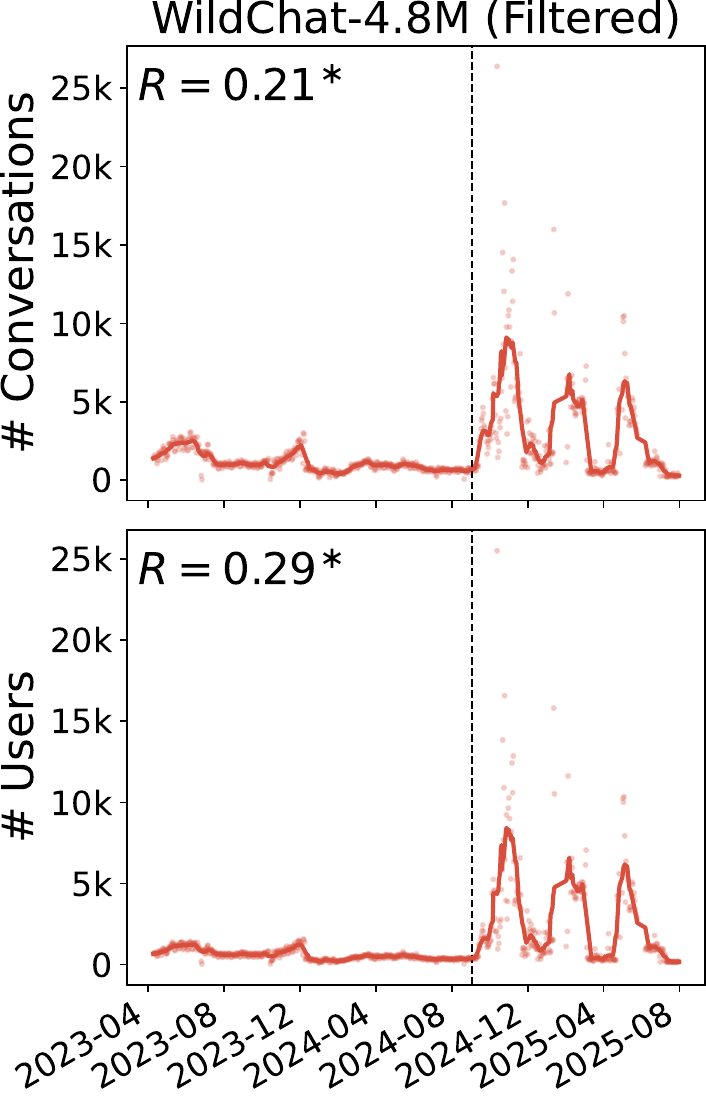}
    \captionof{figure}{Conversation and unique user count in WildChat-4.8, including data after our 2024-09 cutoff (dashed line). Left column: linear $y$-axis, middle column: log $y$-axis, right column: filtered data without ``templated'' conversations. After the cutoff date, there are massive spikes in conversation and user counts, aligning with large increases in the number of API-like templated conversations (see \Cref{fig:templated-metrics} and \Cref{tab:templates}). Filtering out templated conversations reduces the size of these spikes by more than half (right), but there remain an anomalously large number of users and conversations, motivating the use of our temporal cutoff. (Our simple and restrictive definition of templated conversations likely misses other API-like usage.)}
    \label{fig:wildchat-full-counts}
}
\vspace{2em}
{\begin{minipage}[t]{0.495\textwidth}
    \centering
    \includegraphics[width=0.492\linewidth]{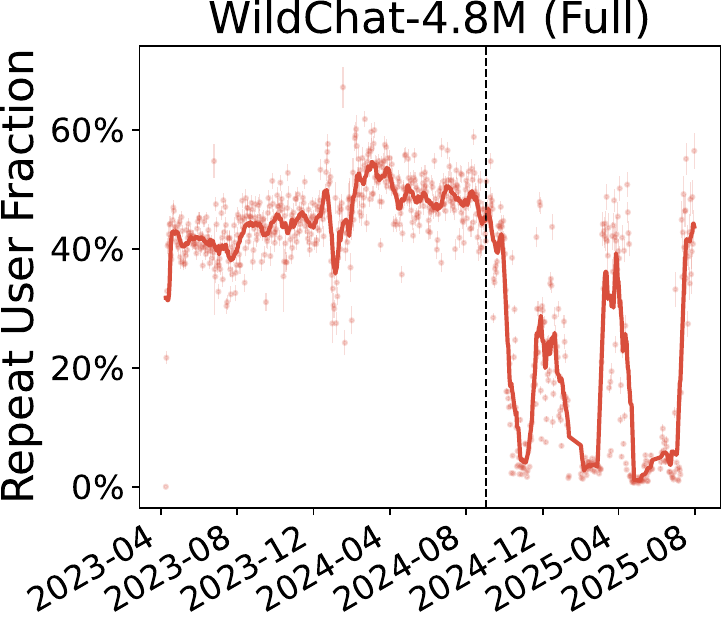}
    \includegraphics[width=0.492\linewidth]{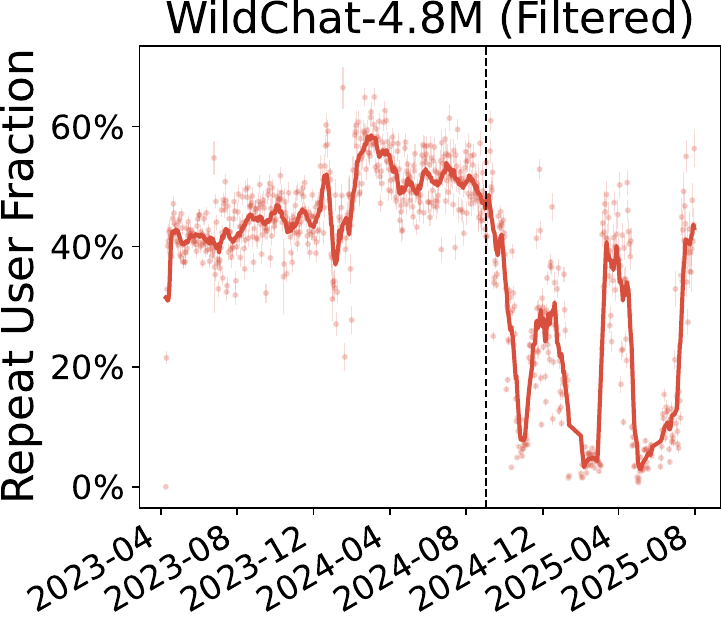}
    \captionof{figure}{Fraction of conversations in each day initiated by a previously-active user. After the cutoff date (dashed line), there are significant drops in repeat users, indicating unusual activity. These coincide with large increases in API-like usage (see \Cref{fig:templated-metrics} and \Cref{tab:templates}).}
    \label{fig:wildchat-full-repeat-user}
\end{minipage}\hfill
\begin{minipage}[t]{0.495\textwidth}
    \centering
    \includegraphics[width=0.492\linewidth]{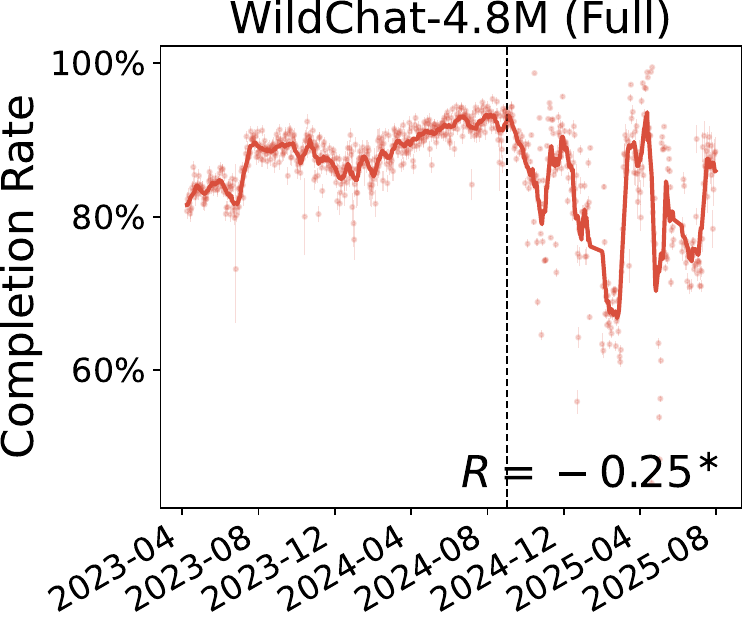}
    \includegraphics[width=0.492\linewidth]{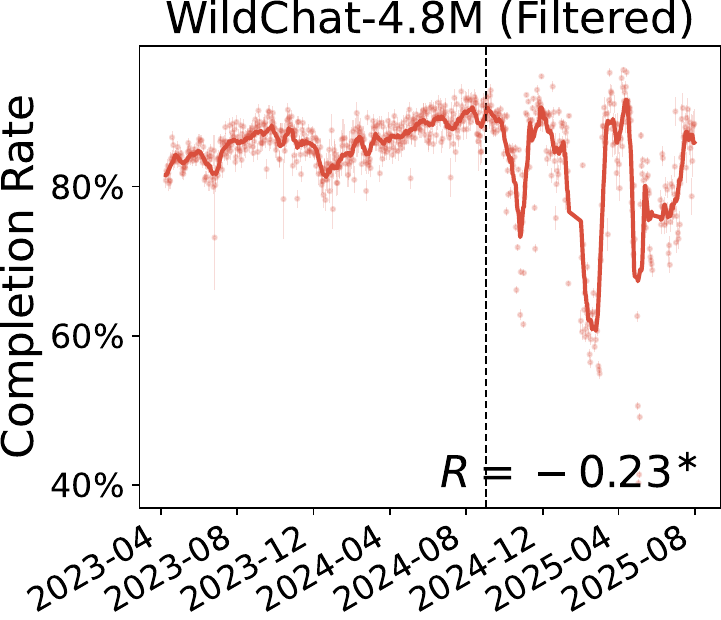}
    \captionof{figure}{Full and filtered versions of \Cref{fig:randomCompletion}, right.}
    \label{fig:wildchat-full-completion}
\end{minipage}}
]

\newpage

\twocolumn[
{
\centering
    \begin{minipage}[b]{0.492\textwidth}
        \centering
        \includegraphics[width=0.492\linewidth]{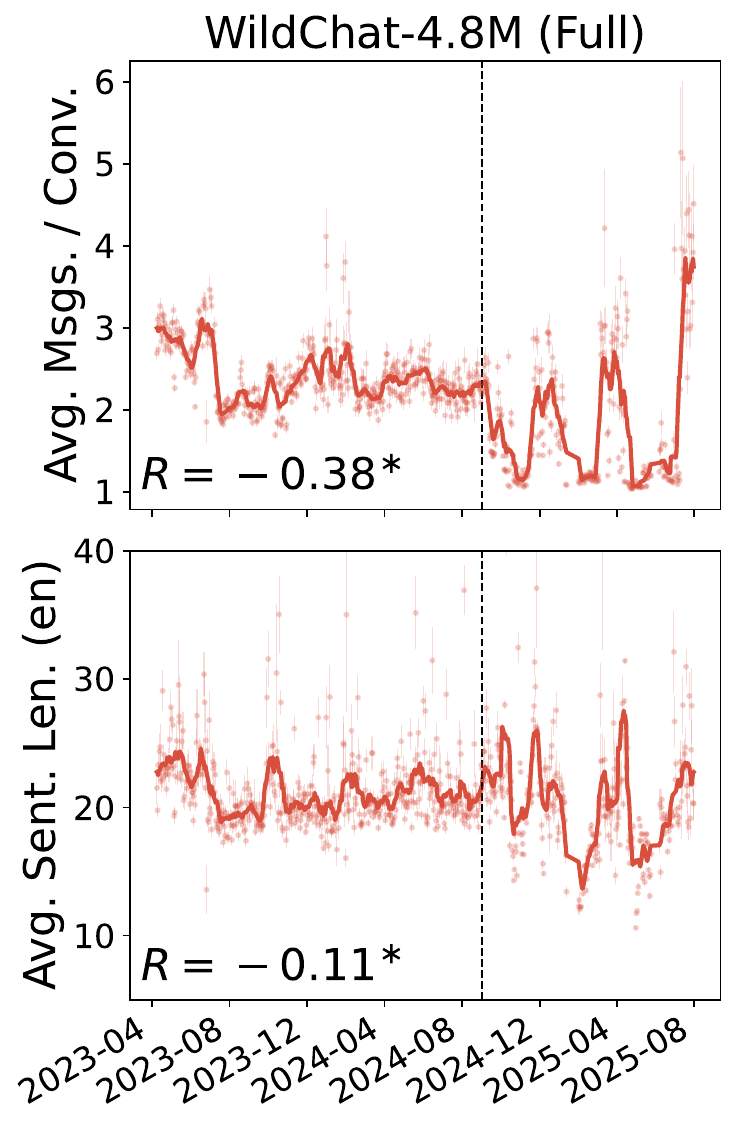}
        \includegraphics[width=0.492\linewidth]{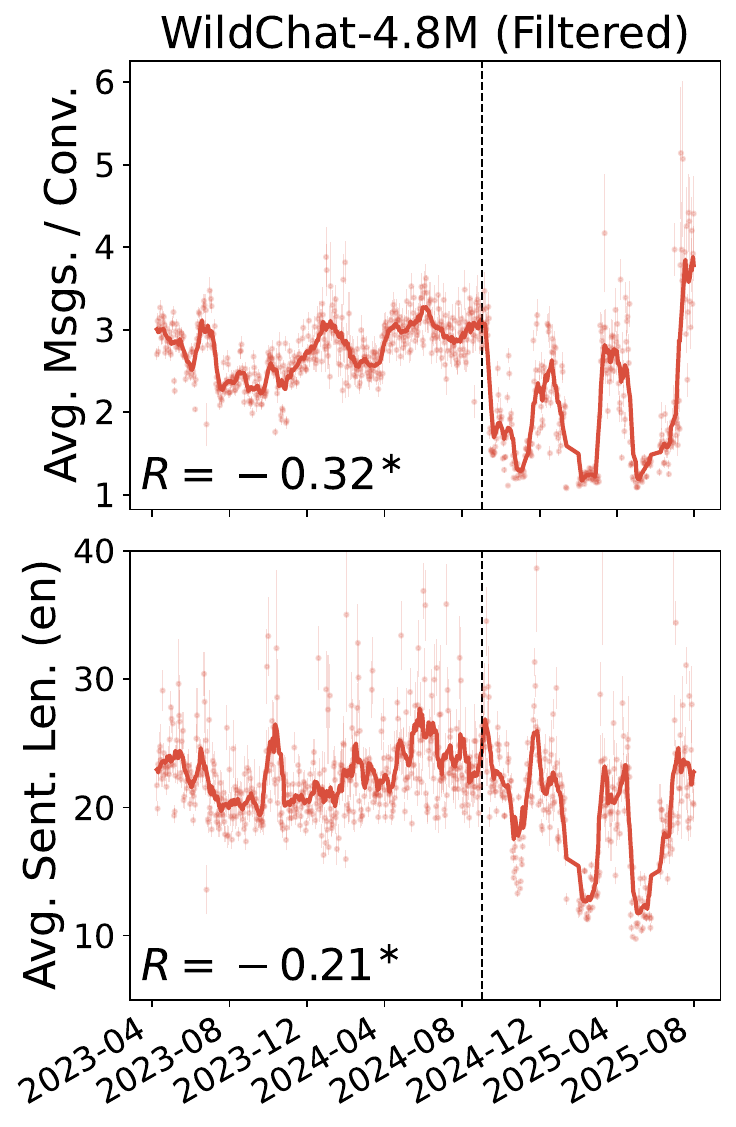}
        \captionof{figure}{Full and filtered versions of \Cref{fig:randomConvInfo}, right.}
        \label{fig:wildchat-full-complexity}
        \vspace{.1em}
        \includegraphics[width=0.492\linewidth]{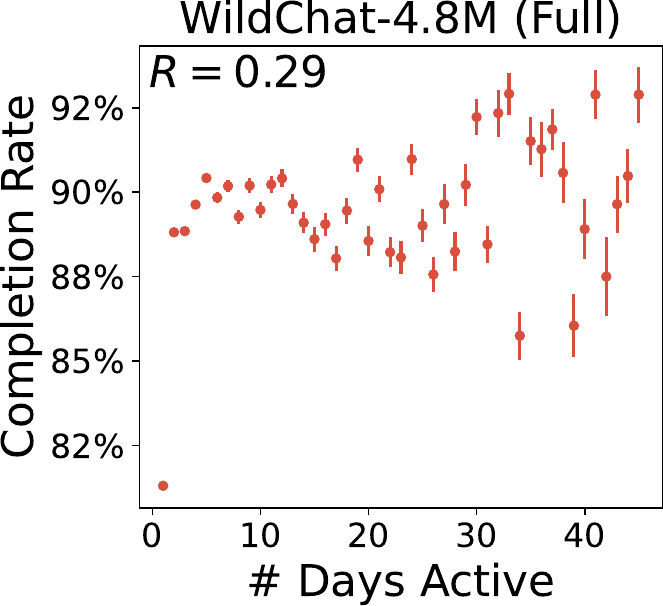}
        \includegraphics[width=0.492\linewidth]{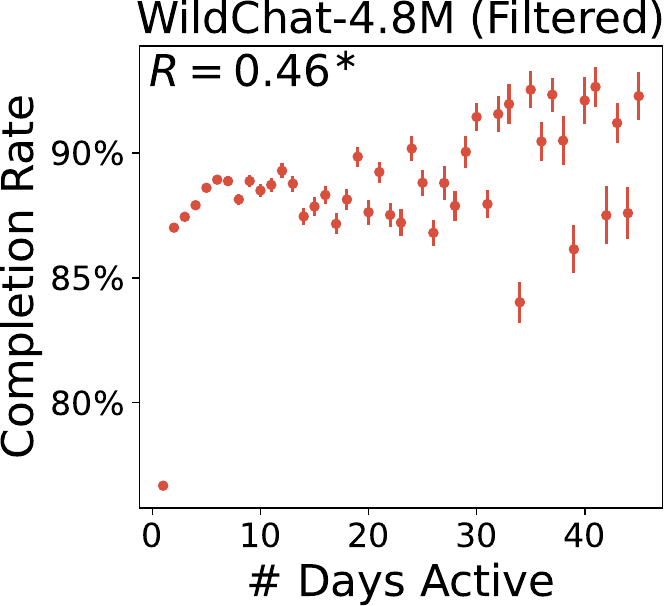}
        \captionof{figure}{Full and filtered versions of \Cref{fig:activeDaysCompleted}, right.}
        \label{fig:wildchat-grouped-data}
    \end{minipage}
    \hfill
    \begin{minipage}[b]{0.492\textwidth}
       \centering
        \includegraphics[width=0.478\linewidth]{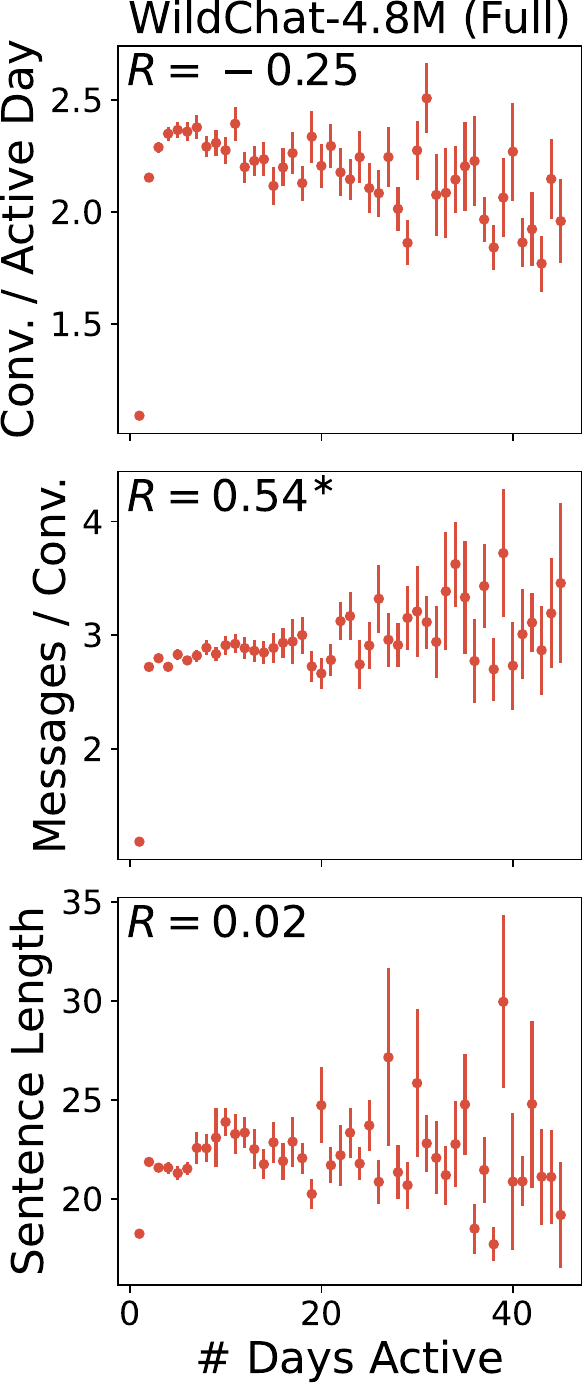}
        \includegraphics[width=0.506\linewidth]{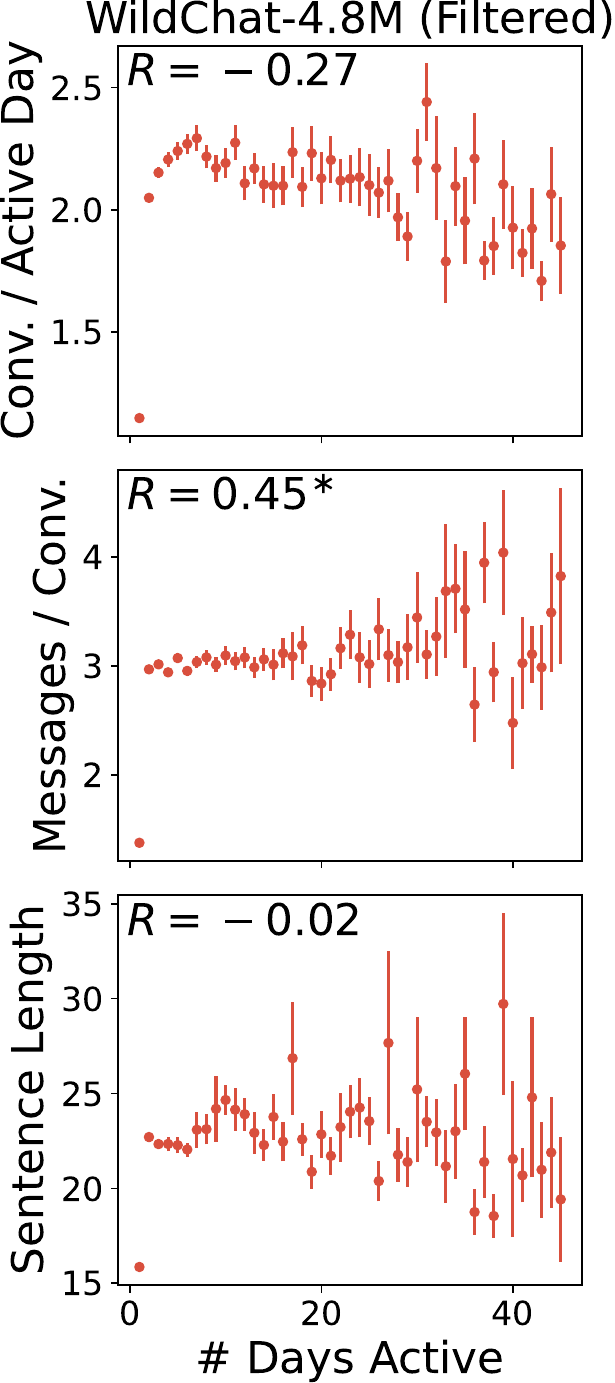}
        \captionof{figure}{Full and filtered versions of \Cref{fig:activeDaysConvInfo}, right.}
        \label{fig:wildchat-full-active-days}
    \end{minipage}
}
\vspace{2em}
{
    \begin{minipage}[t]{0.492\textwidth}
        \centering
        \includegraphics[width=0.492\linewidth]{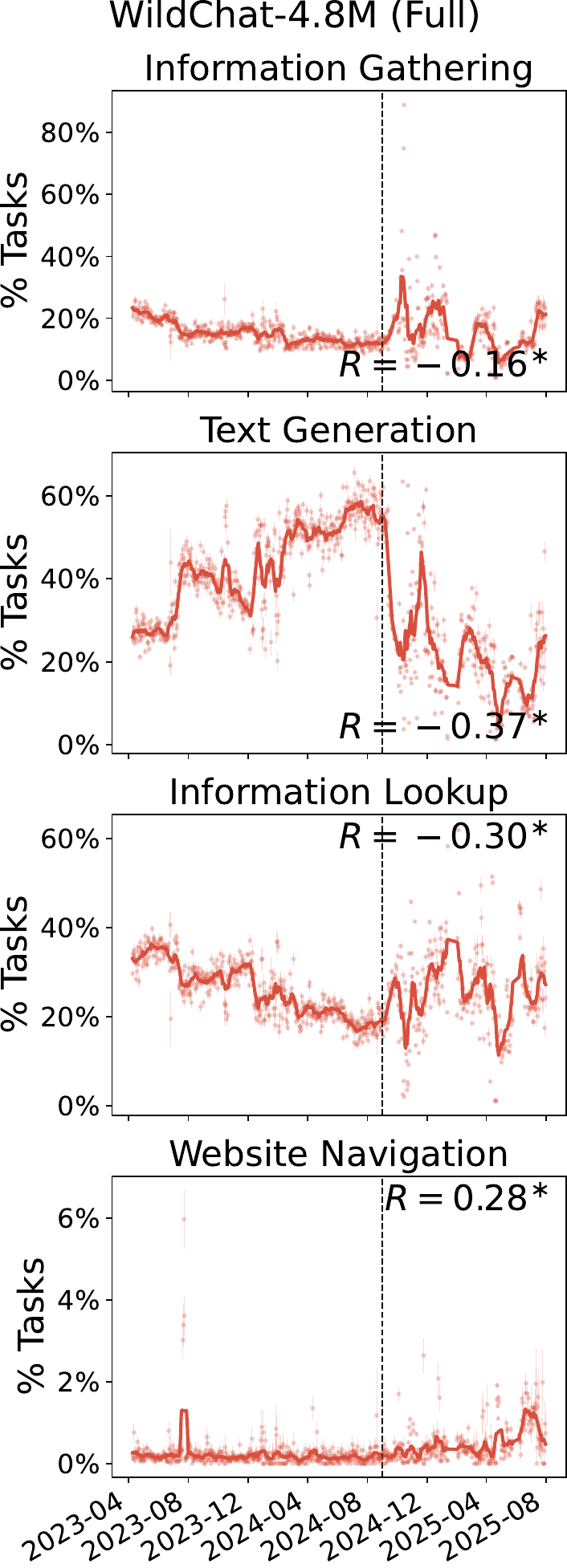}
        \includegraphics[width=0.492\linewidth]{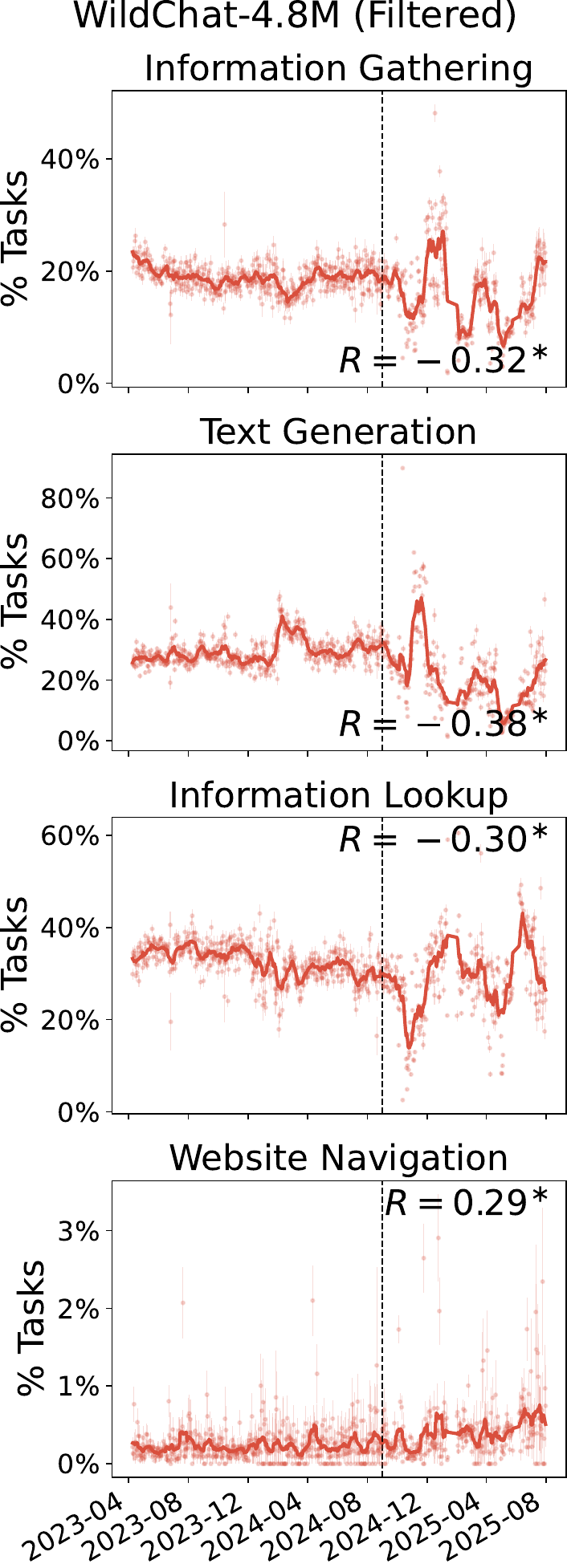}

        \captionof{figure}{Full and filtered versions of \Cref{fig:randomIntents}, right.}
        \label{fig:wildchat-full-feat-intents}
    \end{minipage}%
    \hfill
    \begin{minipage}[t]{0.492\textwidth}
       \centering
        \includegraphics[width=0.492\linewidth]{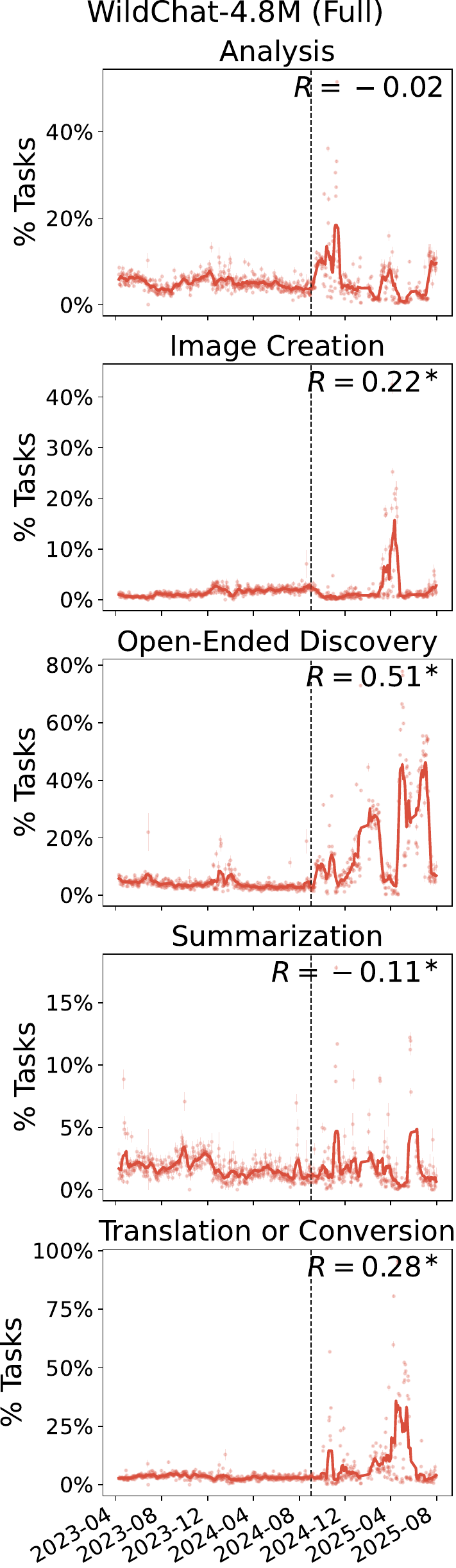}
        \includegraphics[width=0.492\linewidth]{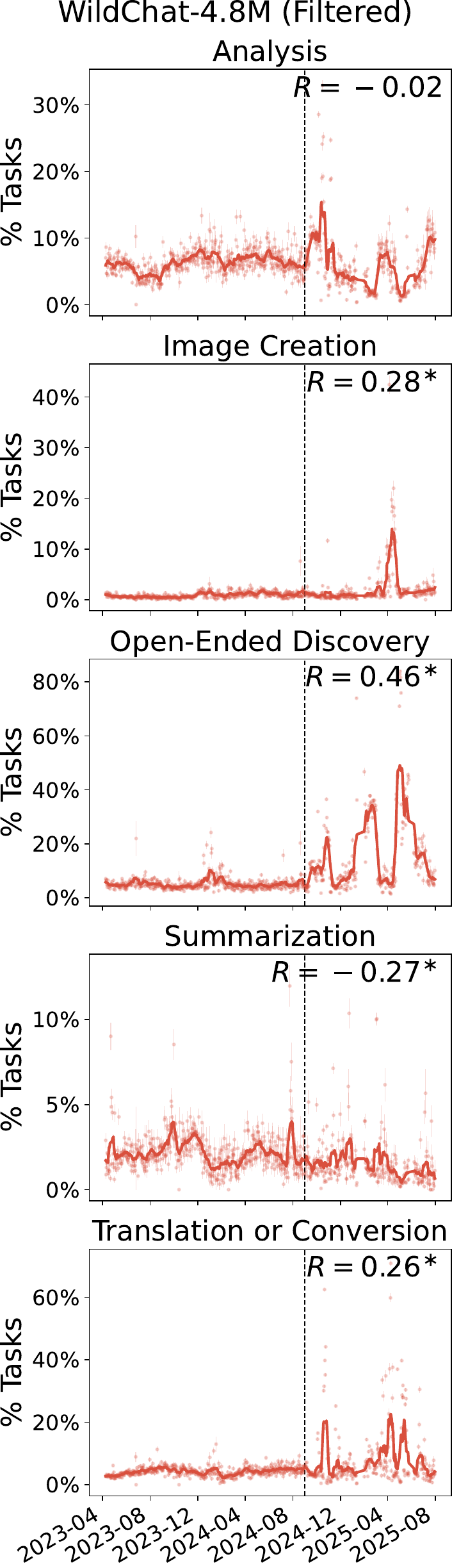}
        
        \captionof{figure}{Full and filtered versions of \Cref{fig:randomIntentsFull}.}
        \label{fig:wildchat-full-intents}
    \end{minipage}
}
]


\begin{figure*}[t]
    \centering
    \includegraphics[width=0.492\linewidth]{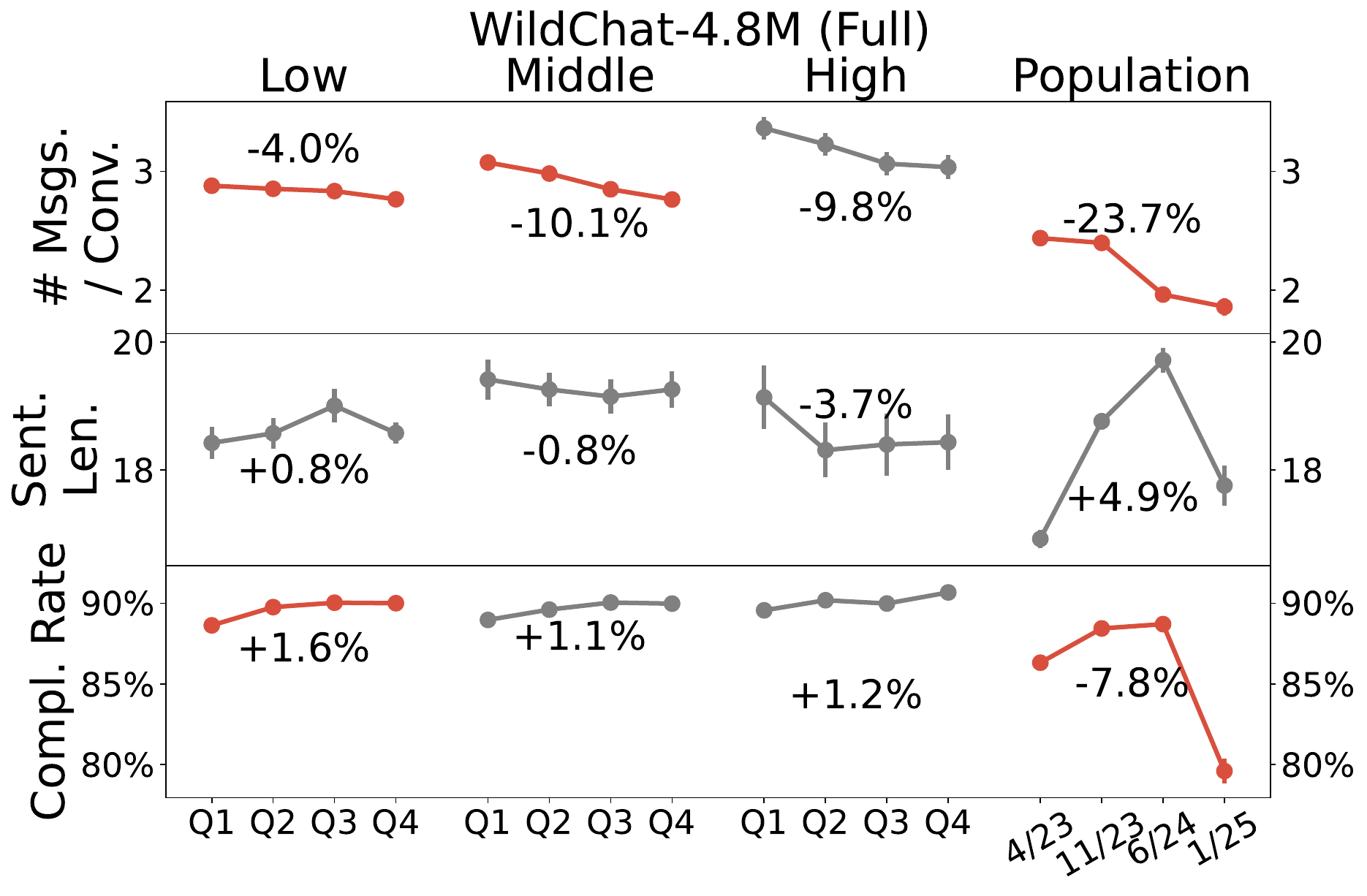}
        \includegraphics[width=0.492\linewidth]{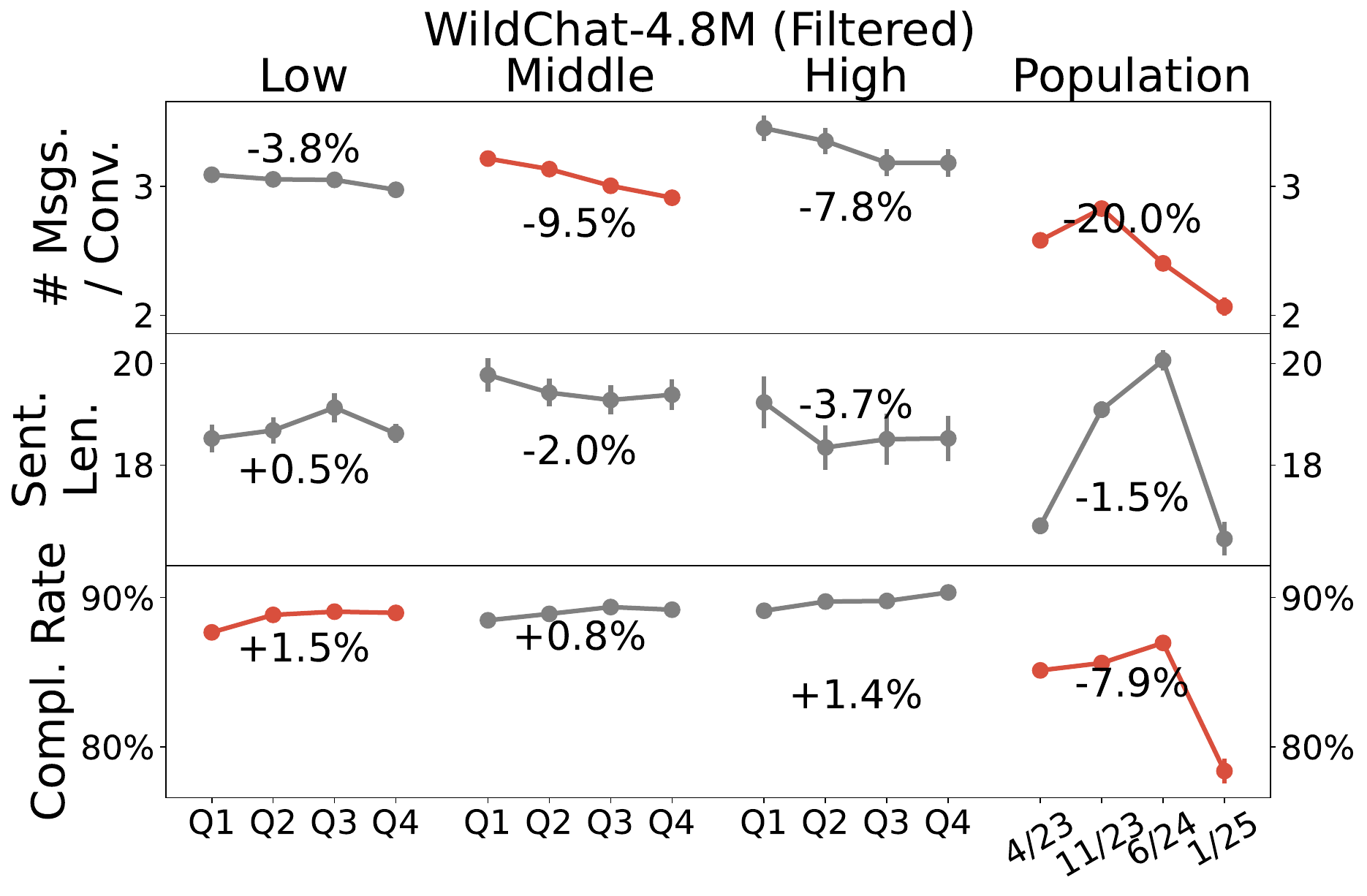}
    \caption{Full and filtered versions of \Cref{fig:activityLevelLifetimeChanges}, right.}
    \label{fig:activityLevelLifetimeChanges-full}
\end{figure*}

\begin{figure*}[t]
    \centering
    \includegraphics[width=0.492\linewidth]{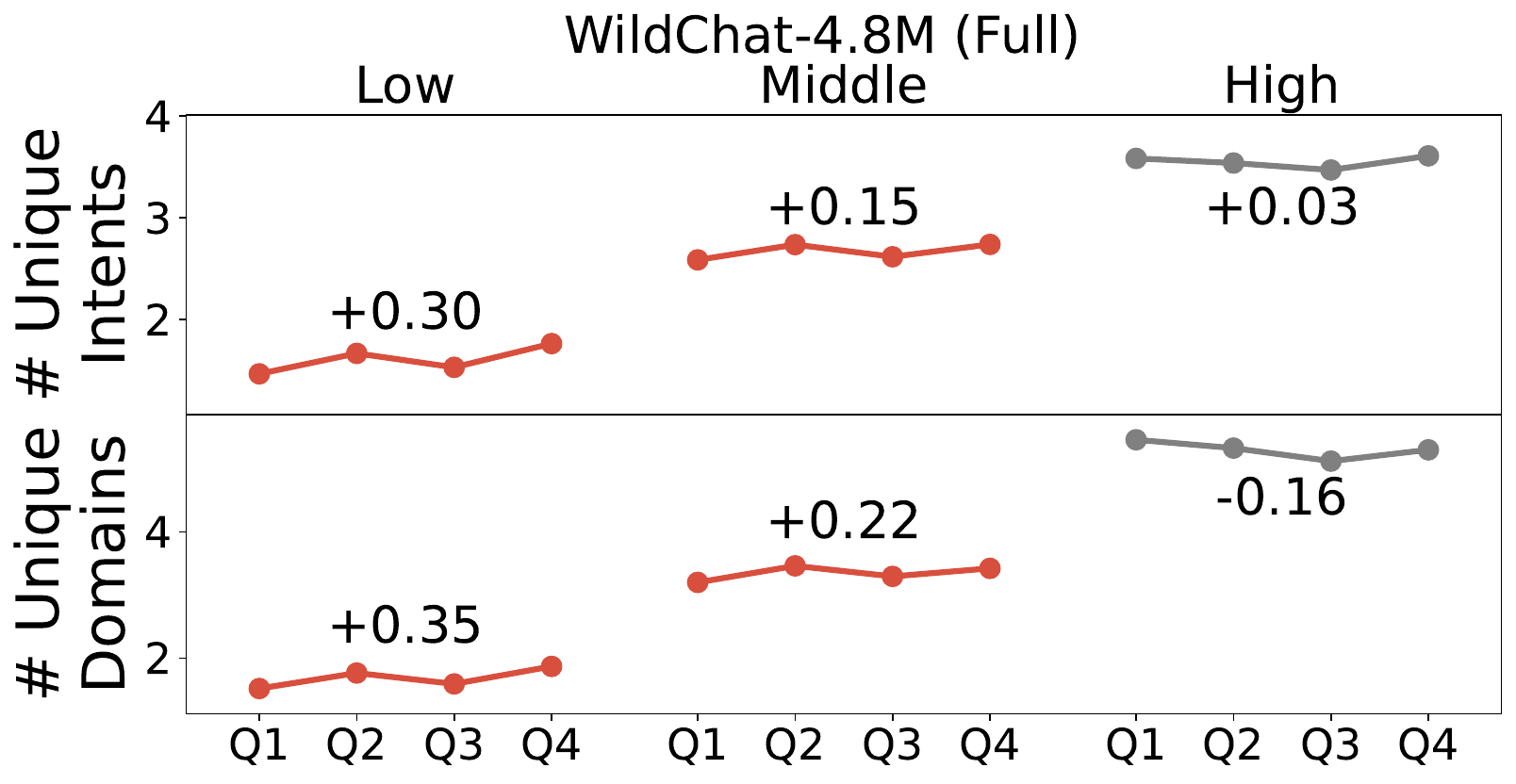}
        \includegraphics[width=0.492\linewidth]{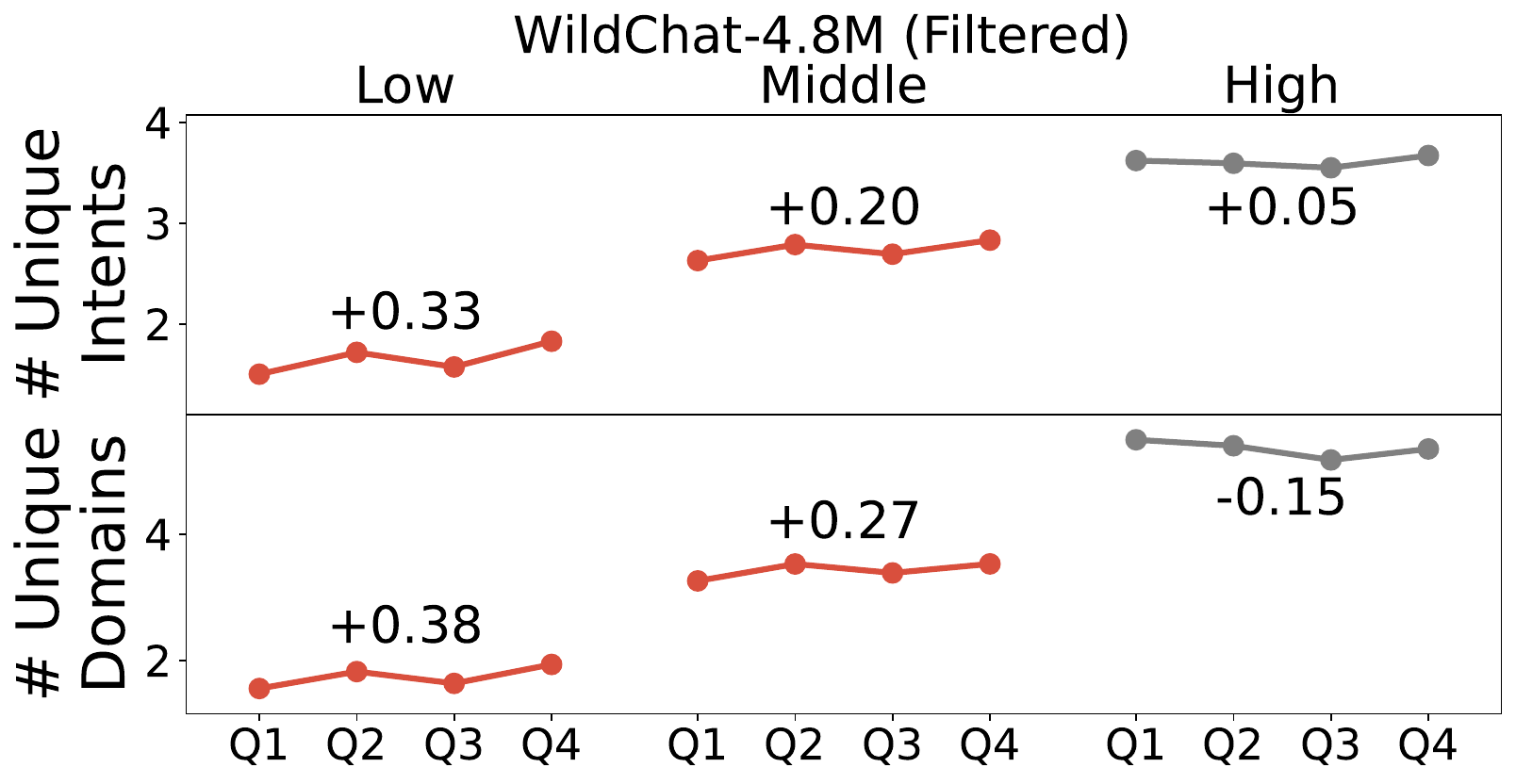}
    \caption{Full and filtered versions of \Cref{fig:numUniqueLifetimes}, right.}
    \label{fig:numUniqueLifetimes-full}
\end{figure*}

\end{document}